%% file: main.tex
\documentclass[10pt]{cai}

\graphicspath{{figs/}}

\usepackage{times}
\usepackage{soul}
\usepackage{url}
\usepackage{graphicx}
\usepackage{amsmath}
\usepackage{amsthm}
\usepackage{booktabs}
\usepackage{algorithm}
\usepackage{algorithmic}
\usepackage{amsfonts}

\usepackage{subfigure}

\usepackage{lipsum}

\usepackage{etoolbox}

\usepackage{textcomp}
\usepackage{xcolor}
\usepackage{xspace}

\usepackage{enumitem}

\newcommand{\squeezeup}{\vspace{-3.39mm}}

\input{macros/macros.tex}

\newcommand{\our}{AppBuddy\xspace}
\newcommand{\R}{\mathbb{R}}

\begin{document}

\begin{center}

  \title{\our: Learning to Accomplish Tasks in Mobile Apps\\ via Reinforcement Learning\footnote{Published in the Proceedings of the 34th Canadian Conference on Artificial Intelligence. Please cite \cite{shvo2021appbuddy} when referencing this paper.}}
  \maketitle

  \thispagestyle{empty}

  \begin{tabular}{cc}
    Maayan Shvo\upstairs{\affilone,*}, Zhiming	Hu\upstairs{\affiltwo}, Rodrigo	Toro Icarte\upstairs{\affilone, *}, Iqbal	Mohomed\upstairs{\affiltwo}, Allan Jepson\upstairs{\affiltwo},
    Sheila A. McIlraith\upstairs{\affilone,*}
  \\[0.25ex]
  {\small \upstairs{\affilone} Department of Computer Science, University of Toronto, Toronto, Canada} \\
  {\small \upstairs{\affiltwo} Samsung AI Center, Toronto, Canada}
  \end{tabular}
  
  \emails{
    \upstairs{*}Work done at Samsung AI Center, Toronto. 
    }
  \vspace*{0.2in}
\end{center}

\begin{abstract}

Human beings, even small children, quickly become adept at figuring out how to use applications on their mobile devices. Learning to use a new app is often achieved via trial-and-error, accelerated by transfer of knowledge from past experiences with like apps. The prospect of building a \textit{smarter} smartphone --- one that can learn how to achieve tasks using mobile apps --- is tantalizing. In this paper we explore the use of Reinforcement Learning (RL) with the goal of advancing this aspiration. We introduce an RL-based framework for learning to accomplish tasks in mobile apps. RL agents 
are provided with states derived from the underlying representation of on-screen elements, and rewards that are based on progress made in the task. Agents can interact with screen elements by tapping or typing. Our experimental results, over a number of mobile apps, show that RL agents can learn to accomplish multi-step tasks, as well as achieve modest generalization across different apps. More generally, we 
develop a platform\footnote{Code, Technical Appendix, and Video Demonstration can be found in \url{https://www.cs.toronto.edu/appbuddy}.} which addresses several engineering challenges to enable an effective RL training environment. Our \our platform is compatible with OpenAI Gym and includes a suite of mobile apps and benchmark tasks that supports a diversity of RL research in the mobile app setting. 
\end{abstract}


\input{sections/intro.tex}
\input{sections/background.tex}

\input{sections/related}

\input{sections/design}
\input{sections/implementation}
\input{sections/evaluation}
\input{sections/discussion}
\input{sections/conclusion}

\printbibliography[heading=subbibintoc]


\newpage

\appendix
\input{sections/supp}

\end{document}

%% file: macros/macros.tex
\newif\ifcomments
\commentstrue

\newcommand{\removehide}[1]{}

\ifcomments

\newcommand{\commentsm}[1]{\textcolor{blue}{(SM: #1)}}
\newcommand{\commentms}[1]{\textcolor{magenta}{(MS: #1)}}

\newcommand{\old}[1]{\textcolor{red}{\sout{#1}}}

\else
 \newcommand{\commentsm}[1]{}
 \newcommand{\commentms}[1]{}
 \newcommand{\old}[1]{}
 
\fi

\newcommand{\tuple}[1]{\langle #1 \rangle}

%% file: sections/intro.tex
\section{Introduction}\label{sec:intro}

Billions of people around the world use mobile apps on a daily basis to accomplish a wide variety of tasks.
%
Building \emph{smarter} smartphones that can learn how to use apps to accomplish tasks 
has the potential to greatly improve app accessibility and user experience. We explore the use of Reinforcement Learning (RL) to advance this aspiration. 

RL has
%
%
been applied in a diversity of simulated environments
with impressive results \cite{lillicrap2015continuous,mnih2015human,silver2016mastering,tessler2017deep}.
%
However, a myriad of challenges can prohibit the application of RL in real-world settings
\cite{dulac2020empirical}. 
Learning to accomplish tasks in mobile apps is one such 
setting due to the usually large action space (the agent can interact with many elements on every screen), sparse rewards, and slow interaction with the environment (i.e., a physical phone or an emulator) that together make the collection of a large number of experience samples both necessary and arduous.

Recent work
proposed to use supervised learning techniques to train computational agents to accomplish tasks in mobile apps \cite{li2020mapping}. 
A shortcoming of this approach is that it requires the creation of large training sets of labeled data. In contrast, RL agents can learn to solve tasks without human supervision by autonomously learning from interacting with mobile apps, and they can potentially learn better solutions than supervised learning approaches, as shown by previous work (e.g., \cite{silver2017mastering}).
Closer to our work are recent efforts that use RL to solve tasks using 
web interfaces \cite{shi2017world,liu2018reinforcement,gur2018learning,jia2019dom} or that test mobile apps \cite{koroglu2018qbe,degott2019learning,pan2020reinforcement}. We discuss these works in Section \ref{sec:related-work}.


In this paper, we explore whether an RL agent can learn policies that consistently solve tasks in real-world mobile phone apps. Our main contributions 
are as follows:
\begin{itemize}[noitemsep]
\item We formulate the app learning task as an RL problem where the state and action space is derived from the phone's internal representation of screen elements and reward is modeled so as to incentivize intermediate task steps, while learning policies that complete tasks.
%
\item We construct a mobile app learning environment that is engineered to collect experiences from multiple emulators simultaneously. The environment is made compatible with OpenAI Gym~\cite{brockman2016openai} to support various RL algorithms. We also build several tools for efficient provisioning of Android emulators, obtaining emulator states, and interacting with the emulators.

\item We experimentally evaluate our RL agent on a suite of benchmarks comprising a number of apps and tasks of varying difficulty. Results (i) demonstrate that RL agents can be successfully trained to accomplish multi-step tasks in mobile apps\removehide{using PPO, a state-of-the-art RL algorithm \cite{schulman2017proximal}}; (ii) expose the impact of design decisions including reward modeling and number of phone emulators used in training;
 and (iii) demonstrate the ability of our approach to generalize to similar tasks in unseen apps. 
\item We develop the \our training platform that includes the aforementioned mobile app learning environment together with  a suite of mobile app-based benchmarks, allowing researchers and practitioners to train RL agents to accomplish tasks using various apps.
\end{itemize}
This paper represents an important step towards endowing smartphones with the ability to learn to accomplish tasks using mobile apps. 
The release of the \our training platform and suite of benchmarks opens the door to further work on this impactful problem by the broader research community.

%% file: sections/background.tex
\section{Preliminaries}\label{app:ppo}

We begin by defining the relevant terminology regarding MDPs and Reinforcement Learning. We then describe the Proximal Policy Optimization algorithm, which we use in our experiments.

\subsection{Reinforcement Learning (RL)}
RL agents learn optimal behaviour by interacting with an environment \cite{sutton2018reinforcement}. The environment is usually modelled as a \emph{Markov Decision Process (MDP)}. An MDP is a tuple $\mathcal{M} = \tuple{S,A,r,p,\gamma}$, where $S$ is a finite set of \emph{states}, $A$ is a finite set of \emph{actions}, $r:S \times A\times S \rightarrow \R$ is the \emph{reward function}, $p(s_{t+1}|s_t,a_t)$ is the \emph{transition probability distribution}, and $\gamma\in(0,1]$ is the \emph{discount factor}. At each time step $t$, the agent is in a state $s_t \in S$ and selects an action $a_t$ according to a \emph{policy} $\pi(\cdot|s_t)$. A policy is a probability distribution over the possible actions given a state. The agent executes action $a_t$ in the environment and, in response, the environment returns the next state $s_{t+1} \sim p(\cdot|s_t,a_t)$ and an immediate reward $r(s_t,a_t,s_{t+1})$. The process then repeats from $s_{t+1}$. The agent's objective is to find an optimal policy $\pi^*$. This is a policy that maximizes the expected discounted future reward from every state $s \in S$.

The value function $v_{\pi}(s)$ is the expected discounted future reward of following policy $\pi$ starting from state $s$. It can be defined recursively as follows:
$$
v_{\pi}(s) =  \sum_{a\in A}{\pi(a|s)\sum_{s'\in S}{p(s'|s,a) \left( r(s,a,s') + \gamma v_{\pi}(s')\right)}}
$$

\subsection{Proximal Policy Optimization (PPO)}
Proximal Policy Optimization (PPO)~\cite{schulman2017proximal} is a policy gradient method that uses a state-approximation technique (usually a deep neural network) with parameters $\theta$ to estimate a policy $\pi_\theta(a|s)$ and its value function $v_\theta(s)$.

PPO then iteratively updates the parameters $\theta$ searching for a better policy (i.e., a policy that collects more reward). To do so, it first collects experiences by running $n$ agents in parallel for some fixed number of steps. Each agent collects experiences by sampling actions from the stochastic policy $\pi_\theta(a|s)$. Then, all those experiences are gathered together and become a training set that PPO uses to improve its current policy $\pi_\theta(a|s)$. This process then repeats.

To update the parameters $\theta$, PPO uses a loss function that considers three terms. The first term is an entropy bonus that discourages $\pi_\theta(\cdot|s)$ from becoming a deterministic policy (which is useful for exploration purposes). The second term is the square error between $v_\theta(s_t)$ and $v_t^{\text{target}}$ for all $s_t$ in the training set. Note that, for each state $s_t$, we can compute an empirical target for its value function estimation using the rewards that the agent collected from $s_t$ on:
$$
v_t^{\text{target}} = r_t + \gamma r_{t+1} + \cdots + \gamma^{k-1} r_{t+k-1} + \gamma^k v_\theta(\cdot|s_{t+k}).
$$
The final term looks to improve the current policy $\pi_\theta(\cdot|s)$. To do so, the key concept is the \emph{advantage estimation} $\hat{A}_t$. Let $a_t$ be the action that an agent selected from state $s_t$ at time $t$, then its advantage estimation is defined as follows $\hat{A}_t = v_t^{\text{target}} - v_\theta(s_t)$. This is the difference between how much empirical reward the agent received by executing $a_t$ from state $s_t$ and how much reward the agent was expecting to get from state $s_t$. Intuitively, if $\hat{A}_t > 0$ (the agent got more reward than expected), PPO will try to increase the probability of selection action $a_t$ in $\pi_\theta(\cdot|s_t)$ (and decrease it otherwise). Concretely, this final term is defined as follows: 
\[
L^{CLIP}(\theta) = {\hat{ \mathbb{E}}}_t{[\min(\rho_t(\theta)\hat{A}_t, clip(\rho_t(\theta), 1 - \epsilon, 1+ \epsilon)\hat{A}_t)]},
\]
where $clip(a,b,c) = min(max(a,b),c)$, $\epsilon$ is usually set to 0.2, and $\rho_t(\theta) = \frac{\pi_{\theta}(a_t|s_t)}{\pi_{\theta_{\text{old}}}(a_t|s_t)}$.
In this case, $\theta_{\text{old}}$ denotes the parameters before the update (i.e., $\pi_{\theta_{\text{old}}}$ is the policy that collected the experiences) and $\theta$ denotes the parameters after the update. The $clip(\cdot)$ function discourages PPO from making large changes to the current policy (which is relevant for theoretical reasons \cite{schulman2015trust}).

%% file: sections/related.tex
\section{Related Work}
\label{sec:related-work}
Related to our work, RL has been applied to learning web-based tasks  \cite{shi2017world,liu2018reinforcement,gur2018learning,jia2019dom}. 
As in our setting, learning to accomplish tasks on the web suffers from sparse reward. In several of these approaches, a Document Object Model (DOM) representation of the current HTML page is used as part of the RL agent state, not unlike our use of view hierarchies. While our work shares some of its motivation with this body of work, the expensive and slow interaction with the mobile app environment, forced a different approach to the problem. 







Also related, is a body of work that has applied RL to testing mobile apps (e.g., \cite{koroglu2019reinforcement,pan2020reinforcement}). Similarly to our work, this body of work leverages the underlying representation of apps to facilitate RL training in the mobile setting, however their endeavor is fundamentally different. They are training and rewarding RL agents to explore and crash apps (e.g., by identifying valid interactions with on-screen elements or maximizing code coverage by reaching novel states) 
for testing purposes, rather than rewarding agents for accomplishing sparsely rewarded multi-step tasks, as we do in this work. 
While \citeauthor{koroglu2019reinforcement} specify concrete test scenarios, their approach uses tabular RL and terminates the learning process when the agent first finds a sequence of actions that satisfies the test scenario \cite{koroglu2019reinforcement}. In contrast, here we leverage deep RL to learn policies that consistently accomplish a variety of tasks in mobile apps.

Recent work by \citeauthor{li2020mapping} proposed to use supervised learning techniques to learn how to map natural language instructions to user interface (UI) elements on the screen of a smartphone \cite{li2020mapping}.  
They adopted crowdsourcing to annotate a dataset coupling natural language instructions (obtained by crawling the web) and corresponding UI elements. Using that training data, they learn a model that maps instructions into sequences of UI elements to interact with in order to complete a task. The model assumes that previous steps have been executed correctly when running multi-step tasks. If, for example, the model selects an incorrect UI element at some point along the sequence, then the next prediction will likely fail because the phone is now in an unexpected state. While our work shares its motivation (and some functionality) with \citeauthor{li2020mapping}'s work, we take a different computational approach, namely RL, that does not require extensive annotations of a large dataset. Rather, our RL agent explores mobile apps and receives rewards from the environments that guide it towards accomplishing tasks.







%% file: sections/design.tex
\section{System Design}~\label{sec:design}
Our objective is to explore whether an RL agent can learn to accomplish tasks in mobile apps by interacting with (either a physical or an emulated) phone environment. Mobile apps are interesting and challenging real-world RL benchmarks. Since they are optimized for accessibility, most tasks can be solved after executing a short sequence of actions. However, the branching factor (i.e., action space) is much larger than in a standard RL benchmark, which leads to a very sparse reward signal. Moreover, interacting with Android emulators is slow. These ingredients make mobile apps a challenging benchmark with interesting structure that, hopefully, RL agents will learn to exploit when solving these problems. 

\begin{figure}[tb]
\centering
	\includegraphics[width=0.65\columnwidth]{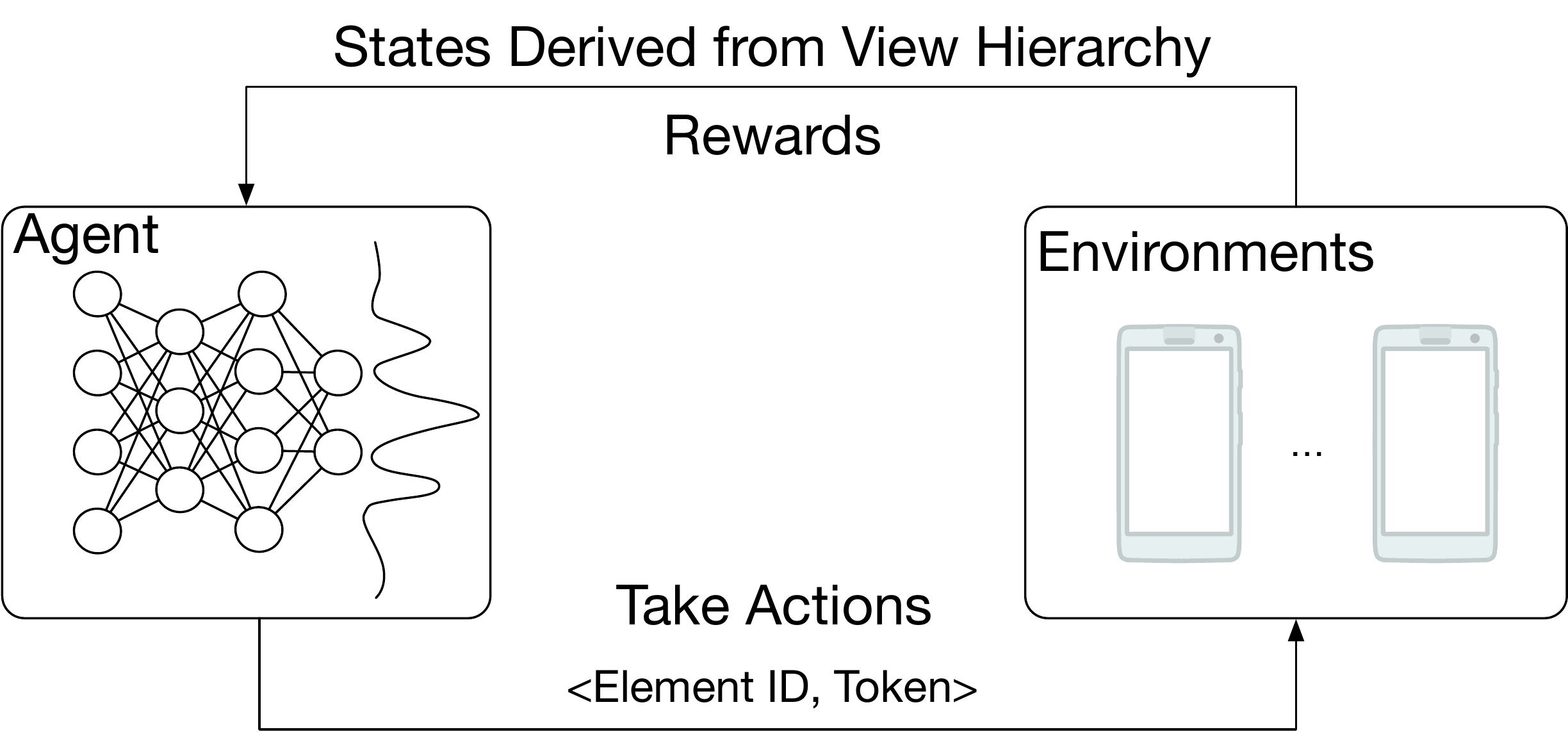}
	\caption{Overview of the proposed RL framework. The environment contains a group of emulators and the states are derived from the view hierarchy of the current screen. In each step, the agent will choose an action, which consists of an element ID and a token, and receive corresponding rewards based on progress made in the task.}
	\label{fig:overview}
\end{figure}

In this section, we present the basic building blocks to use RL in mobile apps: \textit{action space, state representation and reward specification.} The overview of the system is shown in Figure~\ref{fig:overview}. The agent (a neural network trained using PPO) interacts with several Android emulators in parallel to collect experiences. At each step, for each emulator, the agent chooses to tap on (or type into) a particular element on screen. After the agent performs the actions, the environments return \textit{states} derived from the available on-screen information, together with  \textit{rewards} that depend on the current task the agent is solving.

Some tasks might require typing some particular text. For instance, the task ``\textit{add a new Wi-Fi network named Starbucks}" will require the agent to type \textit{Starbucks} at some point. However, it is practically impossible for an RL agent to discover that typing \textit{Starbucks}, letter by letter, will cause it to receive a reward. To handle this issue, we follow the same standard as in the web-based task literature \cite{shi2017world} and provide the agent with a list of tokens to choose from when typing text. With that, we now describe the action space, state representation, and rewards used in our work.



\subsection{Action Space}


The main challenge towards successfully applying RL in smartphones is the unreasonably large action space. Technically, a user might tap on any pixel on the screen. But allowing that level of granularity would make learning infeasible. Fortunately, we can reduce the action space to the set of elements on screen by exploiting the information from the \textit{view hierarchy}\footnote{The view hierarchy is similar to the DOM tree for web pages.}. 

\begin{figure}
\centering
	\includegraphics[width=0.65\columnwidth]{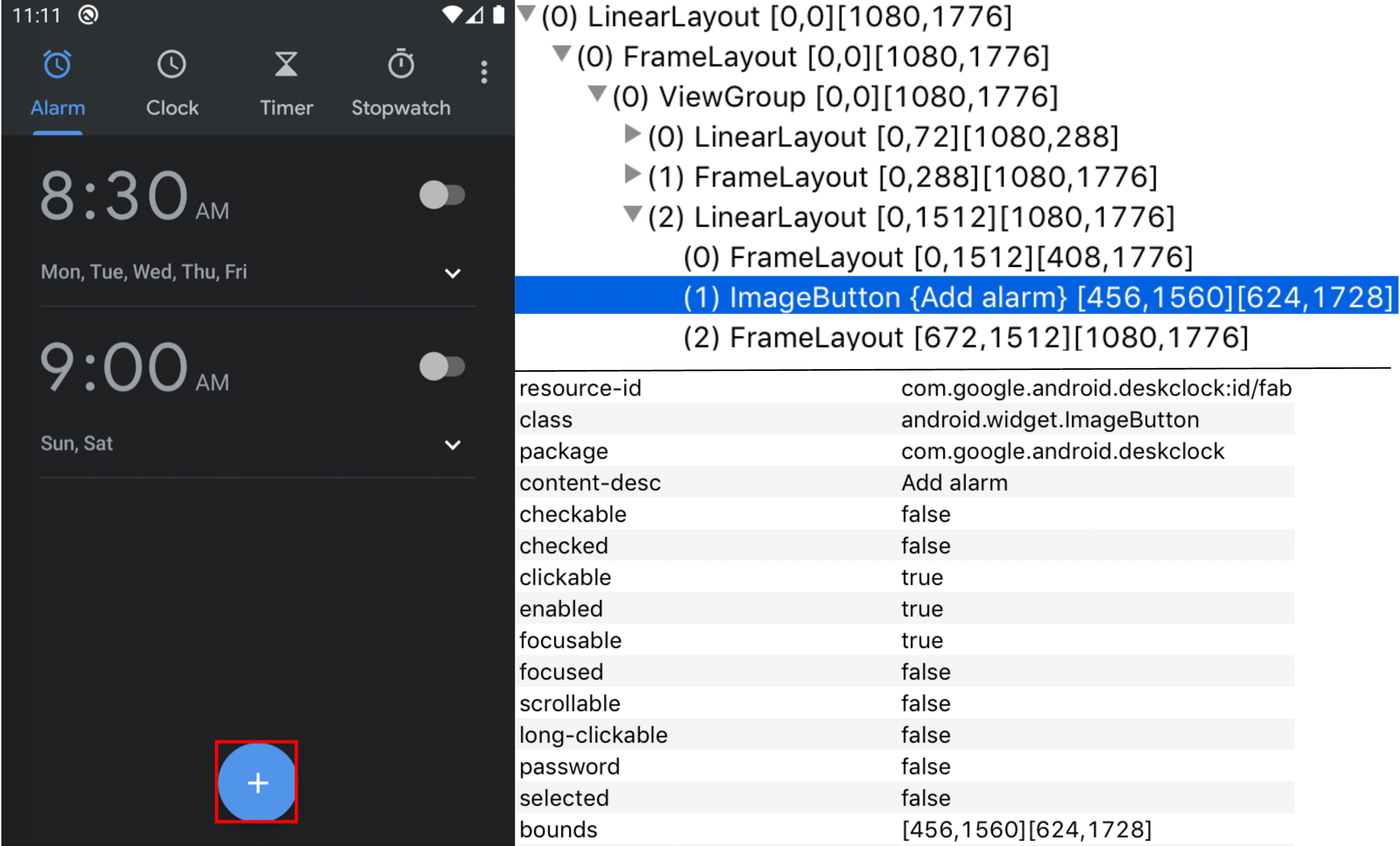}
	\caption{An example of the view hierarchy for a given screen. The `+' button with the red border on the left-hand side directly corresponds to the highlighted element (`Add alarm') in the view hierarchy on the right hand side.}
	\label{fig:view_hierarchy}
\end{figure}

In Android applications, each view is associated with a rectangle on the screen. The view is responsible both for displaying the pixels on the screen as well as handling events in that rectangle. All the views on a particular screen are organized in a hierarchical tree structure, which is also called the view hierarchy. We show a simple example of a view hierarchy in Figure~\ref{fig:view_hierarchy}. In this figure, the left part shows a screenshot of the native Android alarm clock app. On the right, we can see part of the view hierarchy at the top and detailed attributes for the selected view called `ImageButton \{Add alarm\}' at the bottom. The selected view is also highlighted in a red rectangle in the screenshot on the left.

Since the view hierarchy is always available for any Android application, we use its information to define the action space. From this hierarchy, we automatically extract the list of user interface (UI) elements on the screen. Then, the actions at the agent's avail are tuples comprising a UI element index (w.r.t.\ that list) and a token (as shown in Figure~\ref{fig:overview}). The element index ranges from 0 to $n$-$1$, where $n$ is a fixed upper bound for the maximum number of elements on any screen. We say that a UI element is \textit{clickable} if it reacts when tapped and that a UI element is \textit{editable} if text may be typed into it. When the agent chooses to interact with some UI element, the agent will \textit{type} into that element if the element is \textit{editable} and \textit{tap} on that element if the element is \textit{clickable}. 


In addition to selecting a UI element index, the agent also chooses a single token from a set of $k$ tokens (in our experiments, $k$ is 4) that are predefined for each task. For instance, if the task is ``\textit{add a new Wi-Fi network named Starbucks}," then \textit{Starbucks} will be among the $k$ tokens for that task. Then, the chosen token is typed into the selected UI element if that element is editable. Note that it is also possible to train the agent to select correct tokens from natural language commands (e.g., as is done by \citeauthor{jia2019dom} \cite{jia2019dom}). Finally, while we only consider in this work two types of actions -- tapping and typing -- future work will explore action types such as swiping and long tapping.

\subsection{State Representation}
\begin{figure}
\centering
	\includegraphics[width=0.8\columnwidth]{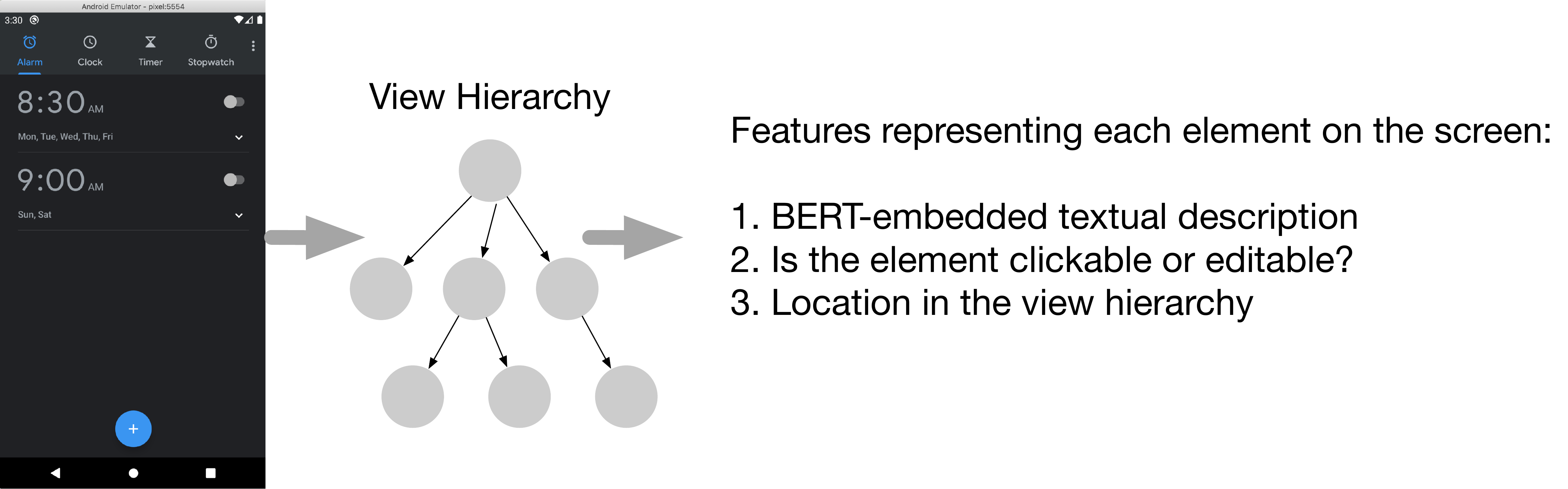}
	\caption{Each element is represented in our state representation by features derived from the view hierarchy of the current screen.}
	\label{fig:state_rep}
\end{figure}

As explained above, the action space is defined by the list of UI elements in the current screen, which is extracted from the screen's view hierarchy. To represent the state, we use the same list of UI elements. Concretely, the state is represented by an $n\times m$ matrix, where each row is a vector of $m$ features representing a particular UI element from the list and $n$ is an upper bound for the maximum number of elements that we expect to see on any screen. In the matrix, each UI element is represented using $m$ features. These features include the \textit{textual description} of the UI element (which is embedded using a pretrained BERT model~\cite{devlin2018bert}). This description is available in the view hierarchy and specifies the purpose of the UI element (e.g., \textit{`Add alarm'} for the `+' button in Figure~\ref{fig:view_hierarchy}). We also include information about whether the UI element is \textit{clickable} or \textit{editable} and its relative location in the view hierarchy (defined as the element's pre-order tree traversal index). The relative location features help capture the spatial correlations across different UI elements. The process of extracting these features from the view hierarchy is illustrated in Figure~\ref{fig:state_rep}. In our experiments, $m$ is 871, i.e., 768 (BERT) + 3 (clickable/editable) + 100 (location in the view hierarchy).


Note that we order the state representation to match the actions in the sense that selecting action $i$ means interacting with the UI element represented by the $i$-th row of the feature matrix. Also, there are cases where the current screen has fewer than $n$ elements. In those cases, we fill the remaining rows of the feature matrix with zeros and, if the agent chooses to interact with nonexistent UI elements, a no-op action is performed (i.e., an action that does not change the environment).



\subsection{Reward Specification}
%

When learning to accomplish tasks in mobile apps, reward is extremely sparse if the agent is only given a positive reward when accomplishing the task and a reward of 0, otherwise.
%
%
%
%
To mitigate for reward sparsity, we specify the reward function $R$ such that the agent is given an intermediate reward for reaching certain states in the app that correspond to sub-goals on the way to accomplishing the task.
%
$R$ is calculated as follows:
\begin{equation}\label{eq:reward_spec}
    R = \sum_{i=0}^{k-1} (w_i \times r_{i}) + w_k \times r_{k} ,
\end{equation}
where $\textbf{w}=\{w_0, w_1,...,w_k\}$ indicates whether the agent has reached a certain state and $\textbf{r}=\{r_0, r_1,...,r_k\}$ represents the rewards. $k$ is the total number of intermediate steps where the agent will receive positive intermediate rewards. The value of $k$ is task-specific. Note that an intermediate reward $r_i$ may only be given once in an episode and $w_i$ is set to 0 after that. 
%
$r_{k}$ is the reward returned by the environment when the task is complete (based on the view hierarchy extracted from the emulator).\footnote{We also tried using reward shaping \cite{ng1999policy} to generate intermediate rewards but the results were inferior.}

For example, in order to accomplish a task in the settings app in our benchmarks, an agent must add a new Wi-Fi network named `Starbucks'. $w_0 = 1$ when the agent reaches the `Wi-Fi settings' screen and $w_1 = 1$ when the agent reaches the `add new Wi-Fi network' screen. $w_2 = 1$ if the agent has added a new Wi-Fi network called `Starbucks' and it now appears on the screen.
In the next section, we discuss our experiments where we show the impact of excluding intermediate rewards (i.e., where only $w_k = 1$ and, hence, the agent receives a sparse reward for accomplishing the task). 

%% file: sections/evaluation.tex
\section{Experimental Evaluation}\label{sec:eval}
%
%
The objectives of our evaluation were: 1) to show that RL can be used to learn policies that accomplish tasks in mobile apps; 2) to expose the impact of design decisions including reward modeling and number of phone emulators used in training; and 3) to demonstrate our approach's potential for generalization to similar tasks in unseen apps.




\begin{table}[]
\centering
\resizebox{1.0\columnwidth}{!}{%
\begin{tabular}{|c|c|c|c|c|c|}
\hline
\textbf{Task}     & \textbf{Steps} & \textbf{Policy Updates} & \textbf{Task}     & \textbf{Steps} & \textbf{Policy Updates} \\ \hline
Settings - Easy   & 1              & 10                     & Alarm - Easy      & 3              & 25                     \\ \hline
Settings - Medium & 2              & 25                     & Alarm - Medium    & 6              & 50                     \\ \hline
Settings - Hard   & 3              & 25                     & Alarm - Hard      & 9              & 75                     \\ \hline
Split - Easy      & 4              & 25                     & Shopping - Easy   & 2              & 25                     \\ \hline
Split - Medium    & 8              & 50                     & Shopping - Medium & 4              & 30                     \\ \hline
Split - Hard      & 13             & 75                     & Shopping - Hard   & 6              & 50                     \\ \hline
\end{tabular}
}
\caption{For each task, \textit{steps} is the minimum number of steps needed to complete the task.}
\label{tab:app_list}
\end{table}

\subsection{Experimental Setup}
%
%
We ran experiments using PPO2 (an implementation of PPO made for GPU \cite{baselines}). The emulators were provisioned using Docker-Android~\cite{docker-android} with headless mode and KVM acceleration enabled. With this setup, we were able to train the agent with tens of emulators on a single machine. 
Below, we provide a high-level description of the domains, hyperparameters, experimental protocol, and evaluation metric. Further details are in Appendix~A.\\












\noindent\textbf{Benchmarks. }
We experiment with 4 mobile apps, where each app includes 3 tasks of varying difficulty. Descriptions of the tasks and the intermediate rewards can be found in Table~\ref{tab:task_descriptions} and Appendix~A, respectively.



\begin{itemize}[noitemsep]
    \item Expense splitting app\footnote{Obtained from \url{https://bit.ly/3bmhhQf}}: create groups of people with whom to split various expenses.
    \item Shopping list app\footnote{Obtained from \url{https://bit.ly/3hVANEr}}: create new checklists, add items to a list, check items off, and remove items from a list.
    \item Alarm clock app\footnote{Obtained from \url{https://bit.ly/3nngfpu}}: add, remove or modify alarms. 
    \item Android networks and Internet settings: add new wireless networks, configure mobile network settings, disable/enable airplane mode.
\end{itemize}
The expense splitting, shopping, and alarm clock apps are open source and come from F-Droid (\url{https://f-droid.org}), which is a repository of open source Android apps, while the settings app comes with the Android OS. Table~\ref{tab:app_list} summarizes the tasks and the minimum number of steps required to accomplish each task.\\

\begin{table}[h!]
\resizebox{\textwidth}{!}{%
\begin{tabular}{|c|l|}
\hline
\textbf{Task}     & \textbf{Task Description}                                                              \\ \hline
Settings - Easy   & Navigate to Wi-Fi settings screen                                                      \\ \hline
Settings - Medium & Navigate to add new Wi-Fi network screen                                               \\ \hline
Settings - Hard   & Navigate to add new Wi-Fi network screen and add a new network called Starbucks        \\ \hline
Split - Easy      & Create a new expense splitting group                                                   \\ \hline
Split - Medium    & Create a new expense splitting group and add a new member to it                        \\ \hline
Split - Hard      & Create a new expense splitting group, add a new member to it, and create a new expense \\ \hline
Alarm - Easy      & Set one alarm clock                                                                    \\ \hline
Alarm - Medium    & Set two alarm clocks                                                                   \\ \hline
Alarm - Hard      & Set three alarm clocks                                                                 \\ \hline
Shopping - Easy   & Add a new item to the default list                                                     \\ \hline
Shopping - Medium & Create a new list                                                                      \\ \hline
Shopping - Hard   & Create a new list and add an item to it                                                \\ \hline
\end{tabular}%
}
\caption{Descriptions of the app-based tasks used in our experiments.}
\label{tab:task_descriptions}
\end{table}

\noindent\textbf{Baselines. }
We show empirically how, by adjusting various knobs, the training process for learning policies to accomplish multi-step tasks in our benchmarks is made easier or harder. More specifically, we compare different configurations of our approach along a number of dimensions.




\begin{itemize}[noitemsep]
    \item Number of emulators: we compare between 3 and 35 emulators (and environments in PPO2).
    \item Reward specification: we compare between a reward specification that includes intermediate rewards and one that does not. For the latter case, $w_0,...,w_{k-1}$ are always set to 0 in Equation~\ref{eq:reward_spec}.
    \item Episode length: we compare between resetting the environment after 25 and 40 steps.
\end{itemize}

In each of the comparisons, only one knob is changed and the rest stay fixed. The `vanilla' configuration includes 35 emulators, intermediate rewards, and an episode length of 25.\\

\noindent\textbf{Experiment Protocol. }
%
In each experiment, we ran PPO2 for some number of policy updates and evaluated the agent's current policy after each update (the number of policy updates per task are listed in Table~\ref{tab:app_list}). To evaluate the policy, we estimated its success rate by running the policy 100 times and counting how many times the policy was able to accomplish the task within 25 steps.

Note that, during training, whenever an episode ends (after 25 or 40 steps or when the task has been accomplished), we reset the emulator by returning the app to a state that is identical to the state of the app immediately following its installation. 
This \textit{hard reset} ensures that the agent always has to solve the task from scratch (and cannot take advantage of any progress made in previous episodes).

\subsection{Results}

%



\begin{figure*}
    \centering
    \subfigure[]{
		\includegraphics[width=0.3\textwidth]{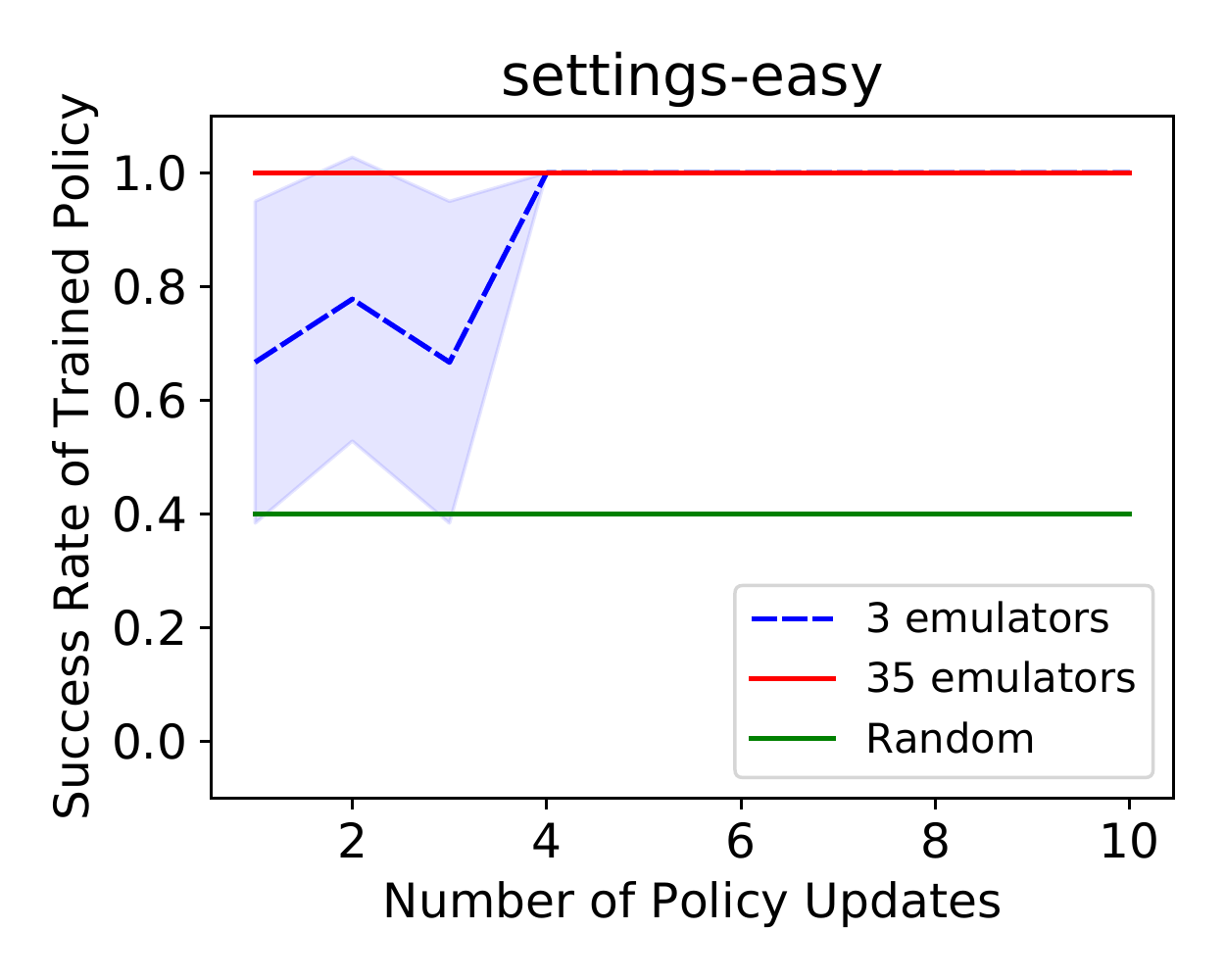}
		\label{fig:settings-easy}
	}
 	\hspace{2pt}
	\subfigure[]{
		\includegraphics[width=0.3\textwidth]{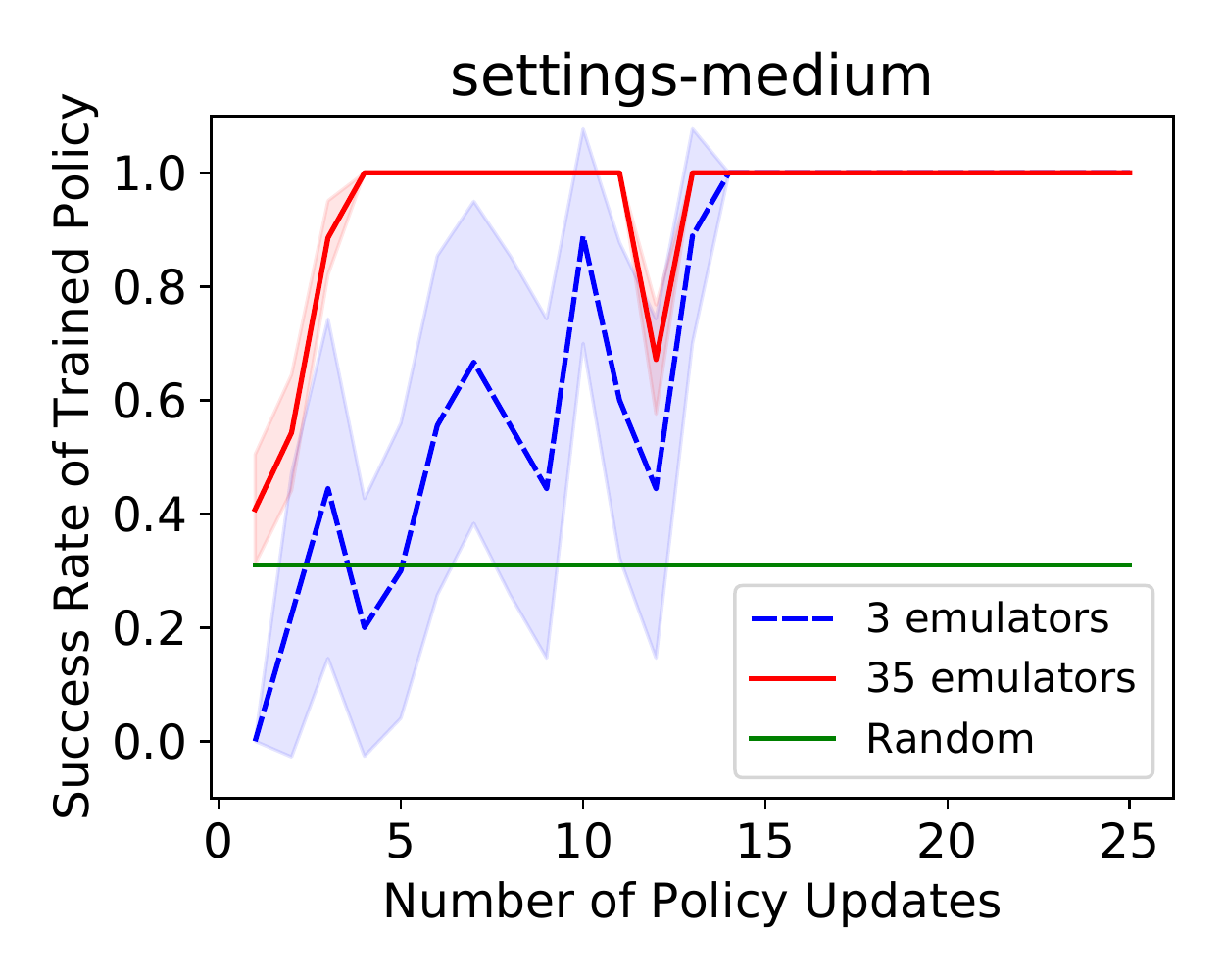}
		\label{fig:settings-medium}
	}
	\hspace{2pt}
	\subfigure[]{
		\includegraphics[width=0.3\textwidth]{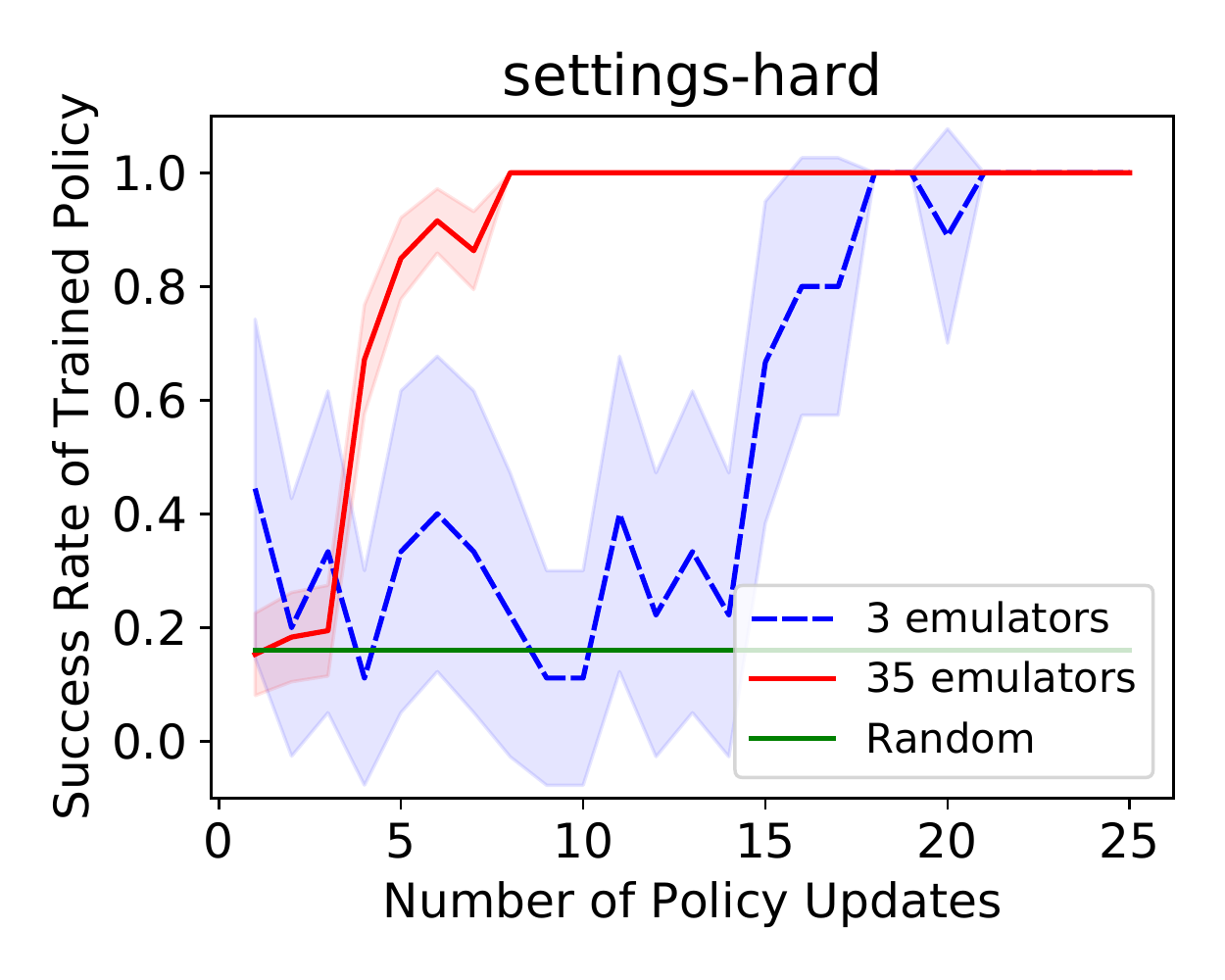}
		\label{fig:settings-hard}
	}
	
	\squeezeup

	\subfigure[]{
		\includegraphics[width=0.3\textwidth]{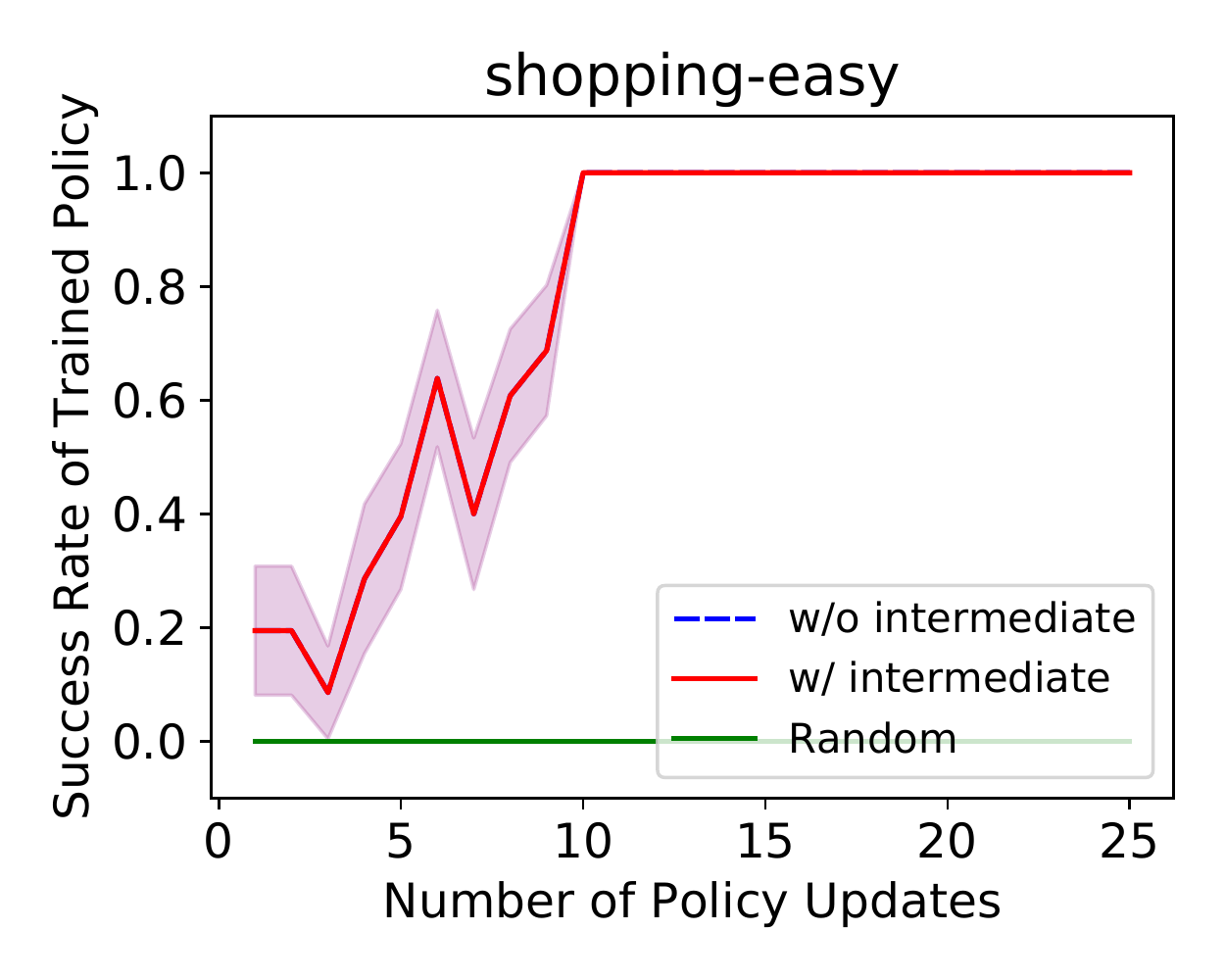}
		\label{fig:shopping-easy}
	}
 	\hspace{2pt}
	\subfigure[]{
		\includegraphics[width=0.3\textwidth]{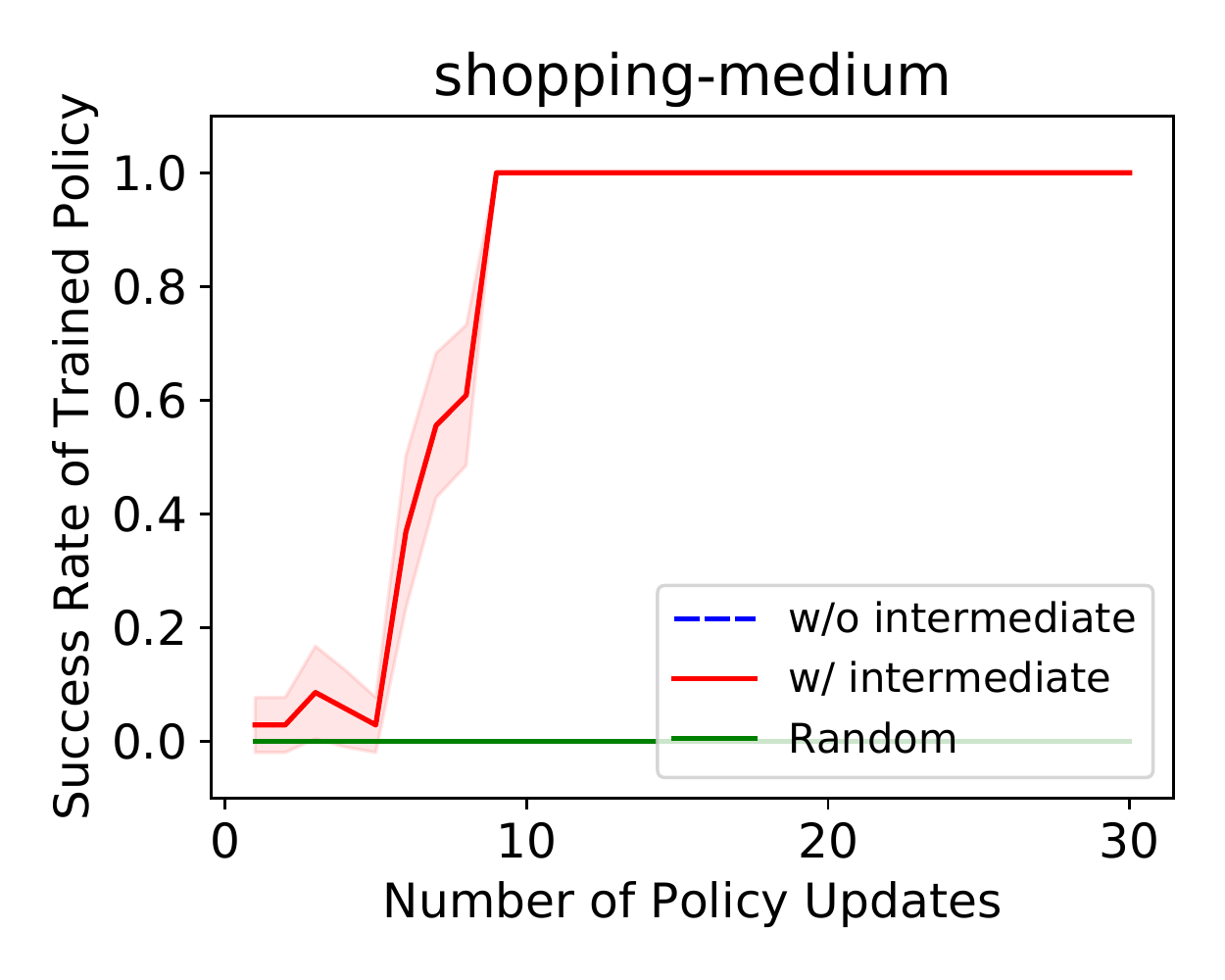}
		\label{fig:shopping-medium}
	}
	\hspace{2pt}
	\subfigure[]{
		\includegraphics[width=0.3\textwidth]{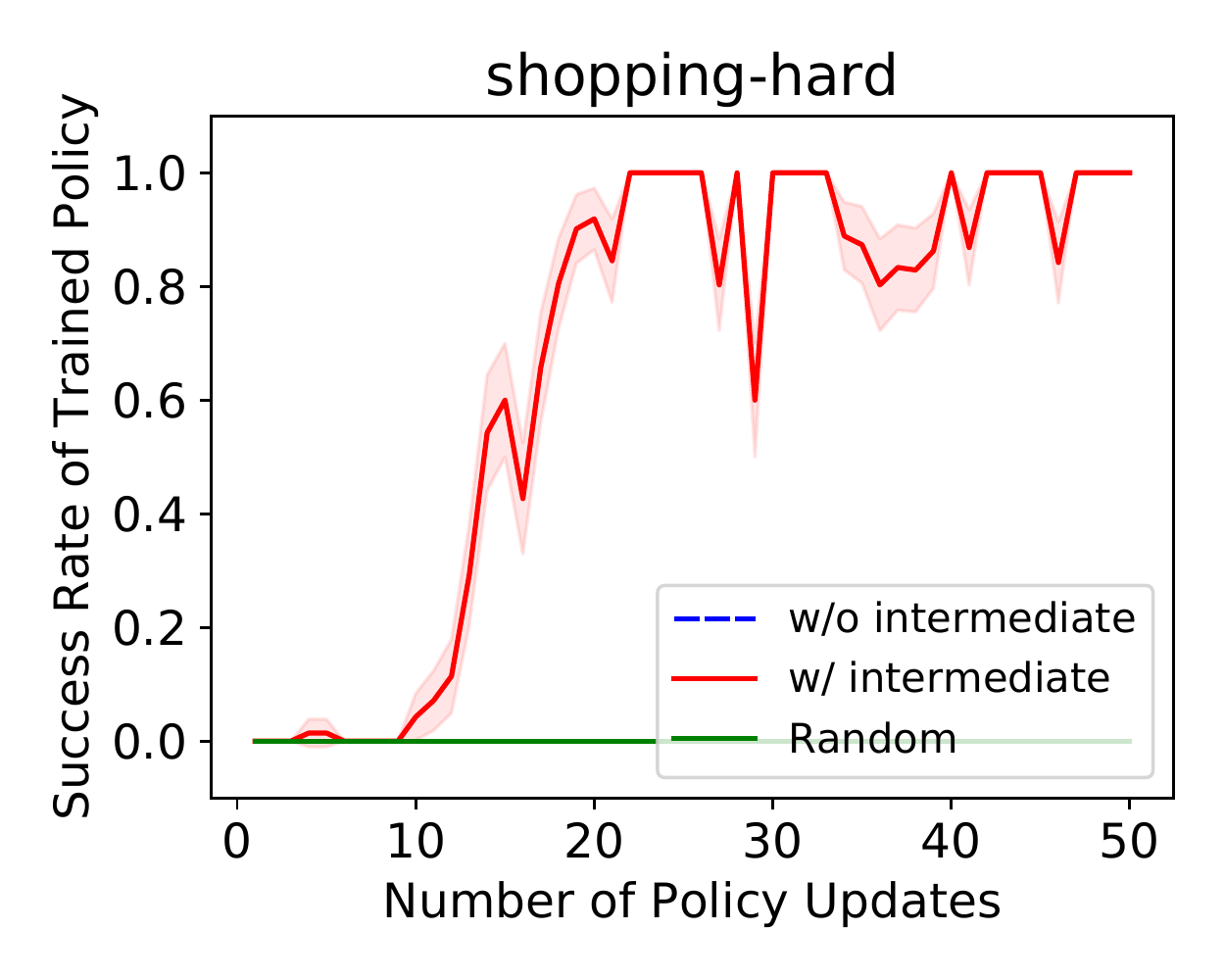}
		\label{fig:shopping-hard}
	}
	
	\squeezeup
	
		\subfigure[]{
		\includegraphics[width=0.3\textwidth]{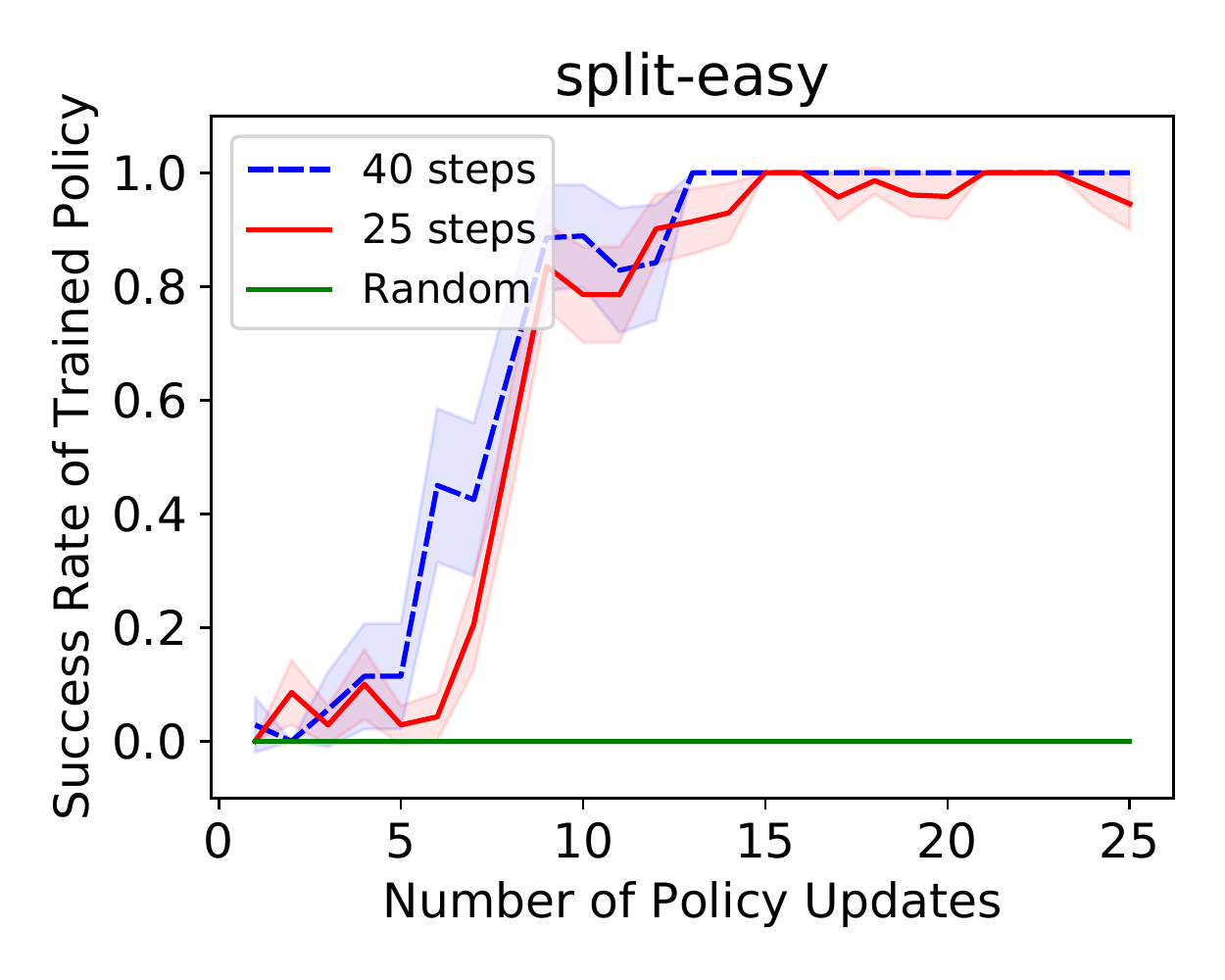}
		\label{fig:split-easy}
	}
 	\hspace{2pt}
	\subfigure[]{
		\includegraphics[width=0.3\textwidth]{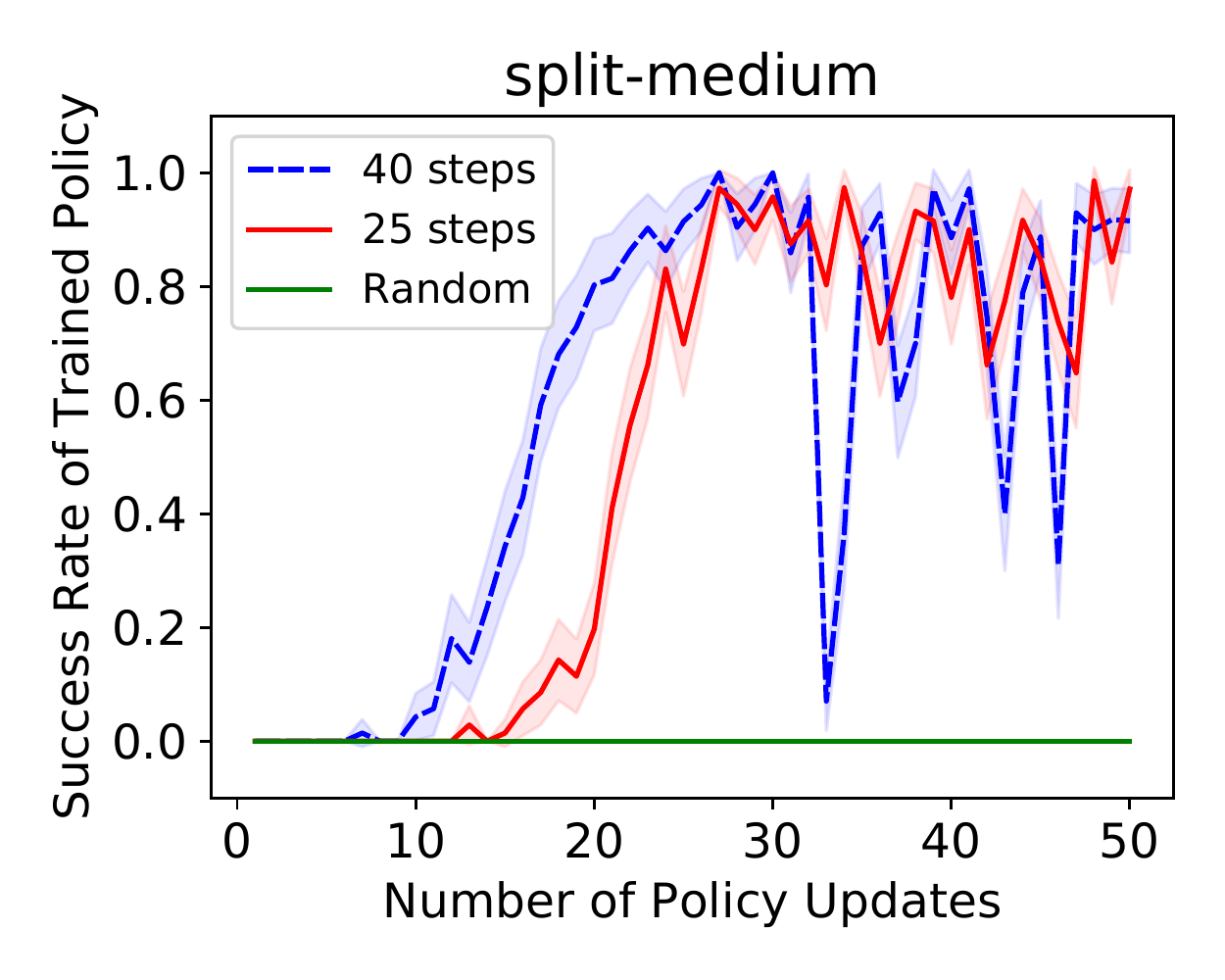}
		\label{fig:split-medium}
	}
	\hspace{2pt}
	\subfigure[]{
		\includegraphics[width=0.3\textwidth]{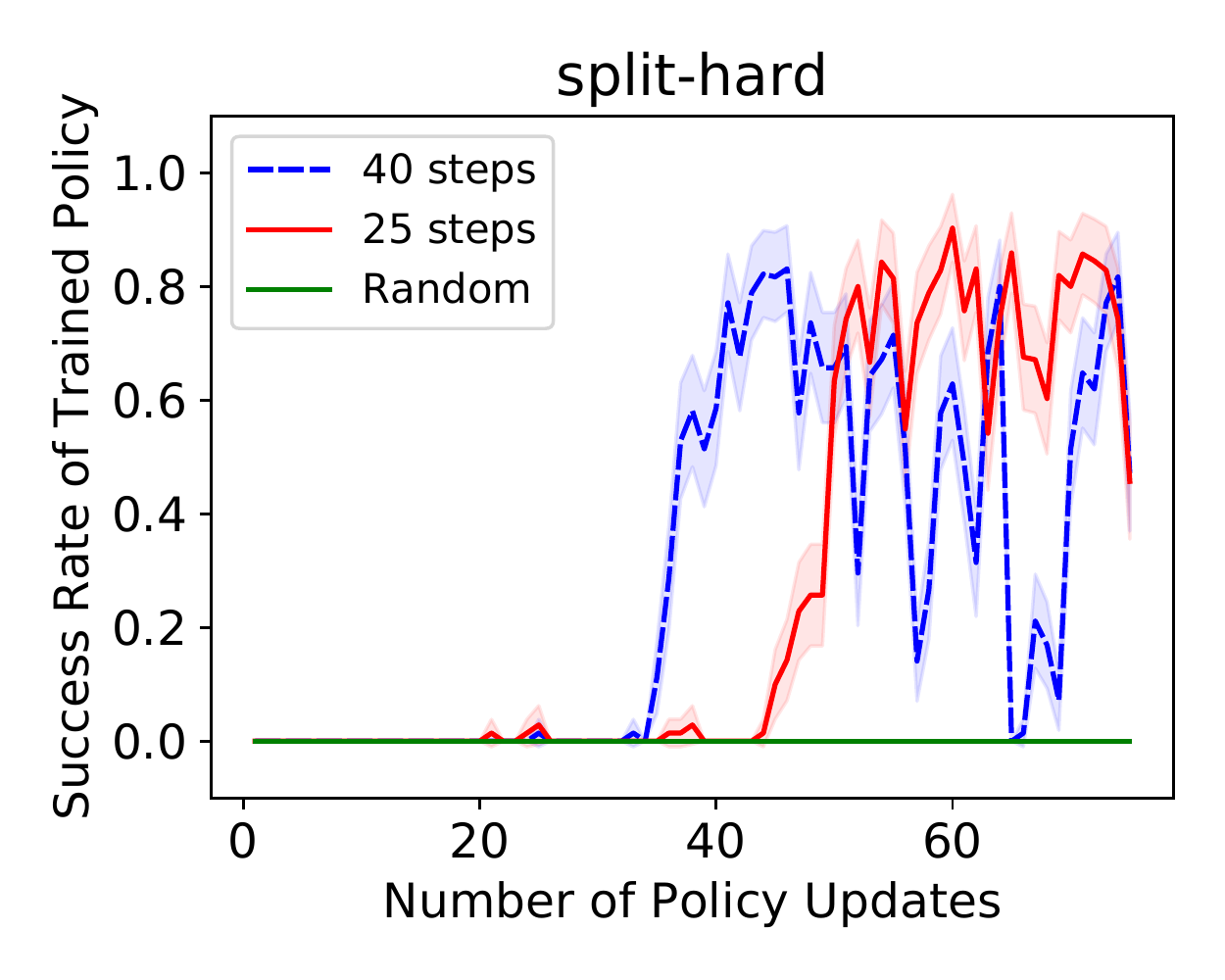}
		\label{fig:split-hard}
	}
	
	\caption{The success rate of a trained policy (percentage of successful test episodes) as a function of the number of policy updates in various tasks. In Figures~(a)-(c), we compare between 3 and 35 emulators. In Figures~(d)-(f), we compare between 25 and 40 steps per episode In Figures~(g)-(i), we compare between a reward specification with and without intermediate rewards. In all the figures, error bars correspond to a 90\% confidence interval. Figures~6-8 (Appendix~B) show the results for the remainder of the tasks.}
	\label{fig:results}
	\squeezeup
\end{figure*}

Figure~\ref{fig:results} shows the success rate of a trained policy as a function of the number of policy updates on various tasks. Each row compares between different baselines. Moreover, we compare to a random baseline that returns a uniformly sampled action at each step. For example, in Figure~\ref{fig:settings-hard}, after 10 policy updates, the trained policy achieves a success rate of approximately 0.1 and 1 with 3 and 35 emulators used in training, respectively. Moreover, in Figure~\ref{fig:settings-hard}, the random baseline achieves a success rate of 0.16. As is evident by Figure~\ref{fig:results}, in most tasks the random baseline was unable to accomplish the task at all.\\

\noindent \textbf{Number of emulators. } Between every two policy updates, 3 emulators gather less experience than 35 emulators. The plots in Figures~\ref{fig:settings-easy}-\ref{fig:settings-hard} and Appendix~B reflect this and show that, on average, the 3 emulator baseline required more policy updates to achieve a high success rate, compared to the 35 emulator baseline. Moreover, in some of the harder tasks (e.g., Figures~6(b),6(c), and 6(f) in Appendix~B), the 3 emulator baseline was unable to accomplish the task at all.\\


\noindent \textbf{Reward specification. }
Figures~\ref{fig:shopping-easy}-\ref{fig:shopping-hard} and Appendix~B show that providing the agent with intermediate rewards was crucial in learning to accomplish many of the tasks. In fact, the baseline that receives no intermediate rewards achieves a success rate of zero in most of the tasks.\\

\noindent \textbf{Episode length. }
The plots in Figures~\ref{fig:split-easy}-\ref{fig:split-hard} and Appendix~B show that in most tasks, the 25-step and 40-step baselines perform similarly. However, there are cases where only the 40-step baseline manages to accomplish the task, such as in Figure~7(h) in Appendix~B.




\subsection{Applying Learned Skills to Unseen Apps}
\label{sec:gen}
Our results indicate that RL can be used to learn policies that accomplish tasks in mobile apps. However, deploying this approach on real phones is still a ways away. In particular, imagine a human user of an alarm clock app. After learning how to use alarm clock app A, the user will typically have an easier time learning how to use alarm clock app B, assuming some similarity between the apps. We would like an RL agent to possess similar capabilities -- learn a policy that accomplishes a task on one app and then use that trained policy to accomplish a similar task on a different, yet similar app.

To experiment with this idea, we begin with a naive approach: we take a policy trained on the easy alarm clock task in our experiments and deploy this trained policy in a different app -- the native Android alarm clock app. The unseen native app has many commonalities to the training app, both in form and in function, however, this naive approach \textit{failed} to accomplish the same task (i.e., setting an alarm clock) in the unseen app, likely because the RL agent did not learn to focus on the appropriate features and instead memorized the element ID and token combination that should be chosen in each state.


To remedy this, we \textit{shuffled} the on-screen UI elements in the state representation given to the agent during training. In this way, the agent can no longer simply memorize the element ID that should be selected and must, instead, learn and attend to the relevant features of each element in the state representation (e.g., the BERT-embedded textual description). However, this approach also failed because the information encoded by the view hierarchy for each element was not sufficiently similar between the two apps. For example, while the accessibility text accompanying the `+' button in the native alarm clock app reads \textit{add alarm} (see the blue button in Figure~\ref{fig:view_hierarchy} and the value of the \textit{content-desc} attribute in the view hierarchy), there is no accessibility text accompanying the corresponding `+' button in the open source app. This discrepancy is due to the developers of each app. We hypothesized that augmenting the accessibility text for the relevant elements in the open source alarm clock app (on which the agent is trained) with text that is similar to the accessibility text accompanying the corresponding elements of the unseen app, will help the agent generalize to the unseen app.\\

\begin{figure*}
    \centering
    \subfigure[]{
		\includegraphics[width=0.4\textwidth]{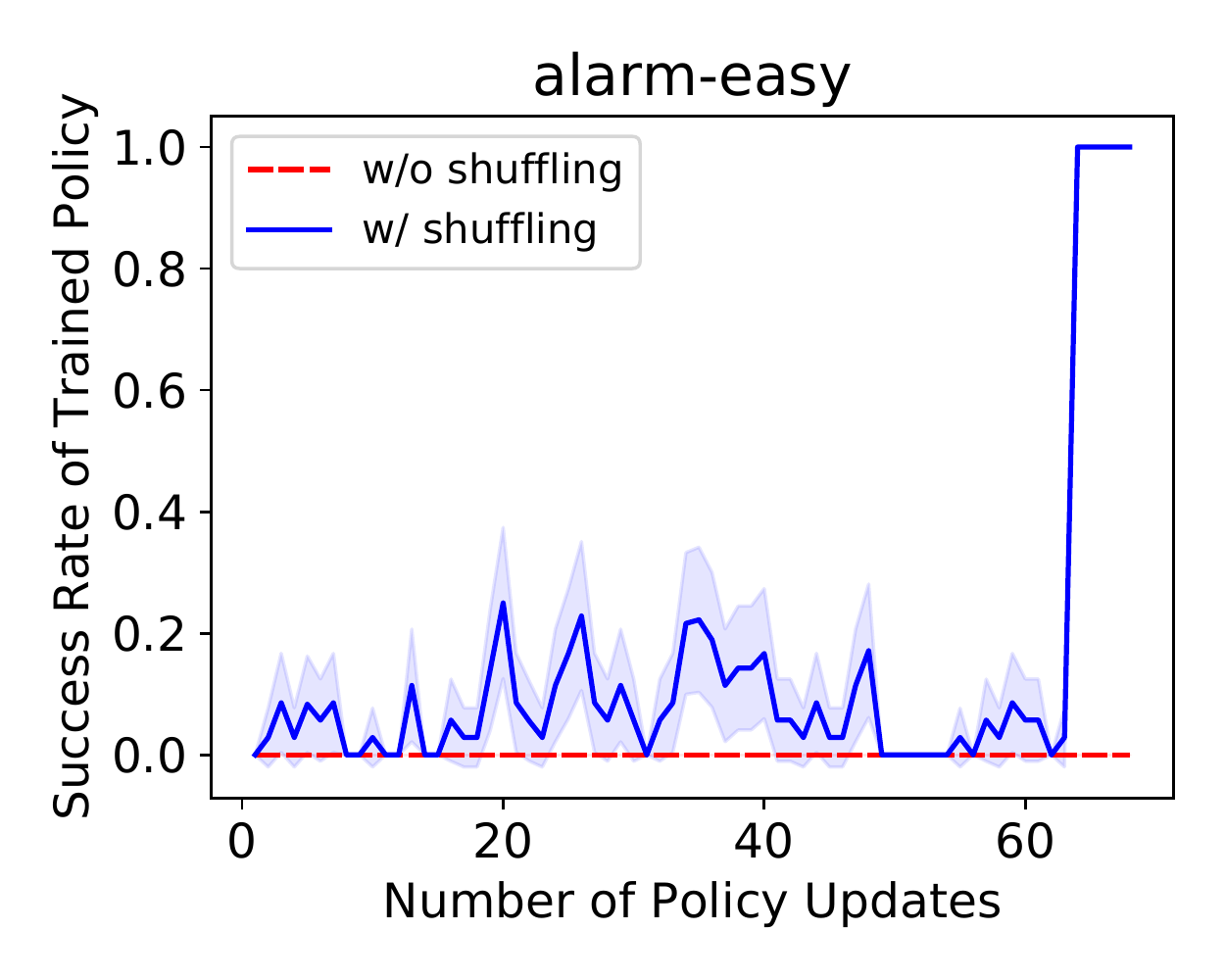}
		\label{fig:shufflingVsNoShuffling_unseenapp}
	}
 	\hspace{2pt}
	\subfigure[]{
		\includegraphics[width=0.4\textwidth]{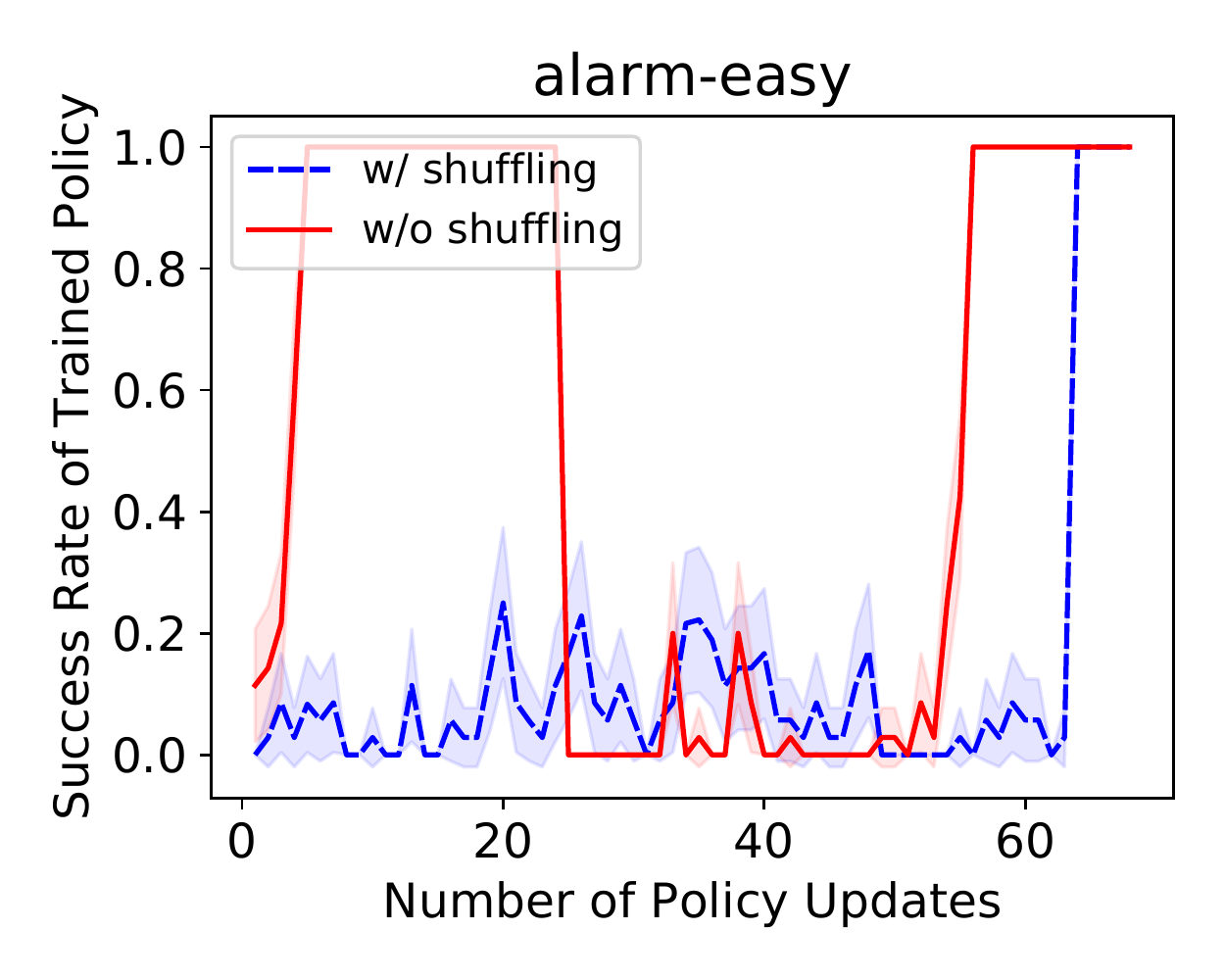}
		\label{fig:shufflingVsNoShuffling}
	}
	
	\caption{The success rate of a trained policy (percentage of successful test episodes) as a function of the number of policy updates. The plots compare between the baseline used in our experiments (i.e., w/o shuffling) and a baseline where the state representation given to the agent is shuffled and augmented with relevant text, on the \textbf{unseen, test app (left)} and on the \textbf{training app (right)}.}
	\label{fig:generalization}
	\squeezeup
\end{figure*}

\noindent \textbf{Generalization Experiments. }
%
To test our hypothesis, we trained a policy on the open source alarm clock app while also shuffling the state representation, which now included relevant text in the accessibility field, as described above.
We compared between the baseline used in our experiments (without shuffling the state representation) and a baseline where the state representation is shuffled and augmented with relevant text.
Figure~\ref{fig:shufflingVsNoShuffling_unseenapp} compares between these two baselines on the unseen app in terms of success rate achieved as a function of the number of policy updates.
Figure~\ref{fig:shufflingVsNoShuffling} compares between these two baselines on the training app in terms of success rate achieved as a function of the number of policy updates. Figure~\ref{fig:shufflingVsNoShuffling} shows that training an RL agent in the training app without shuffling, unsurprisingly, achieves a high success rate after fewer policy updates, compared to an RL agent that is given a shuffled state representation. However, as shown by Figure~\ref{fig:shufflingVsNoShuffling_unseenapp}, only by shuffling the state representation and forcing the agent to pay attention to the features, it was able to generalize to a similar task in the unseen alarm clock app.

\subsection{Discussion}

%

\noindent \textbf{Reward specification. }
In our experiments, PPO2 needed intermediate rewards in order to solve most of our tasks. This is unfortunate because it increases the complexity of programming reward functions. We encourage future work to explore how to learn policies without using intermediate rewards or how to generate intermediate rewards automatically. For example, previous work has proposed to learn structured representations of reward functions from experience and has shown that these representations can be used to effectively solve partially observable RL problems \cite{toro2019learning}.\\

\noindent \textbf{Training efficiency. } 
Interacting with Android emulators is slow and it will be important to consider how to speed up such interactions. Interestingly, the bottleneck in our training was the resetting of the environment - the emulator. In the presence of multiple emulators running on the same machine, parallel resets often caused issues (e.g., the machine was temporarily unable to query an emulator to obtain the current view hierarchy and derive the current state from it). To mitigate for this, we cached the initial view hierarchy of each app, which allowed us to avoid querying the emulator during the reset phase.\\

\noindent \textbf{Partial observability. }
Our state representation is derived from the information available from the current screen. However, that information alone might not be enough to determine the optimal action. For instance, the task ``\textit{add one alarm at 7 am and another at 7 pm}" (our alarm-medium task) becomes partially observable because, when the agent is in the screen for setting a new alarm, it does not have information about whether it had already set the other alarm. As such, providing some form of memory to the agent will be key to solving harder tasks in mobile apps.\\
%
%

\noindent \textbf{Generalization.}
Our results on generalization (Section~\ref{sec:gen}) show that RL agents can learn policies that generalize to unseen (but similar) apps. However, we had to manually add meaningful textual descriptions to the view hierarchy to do so. Fortunately, adding such descriptions can be done automatically by using a widget captioning technique~\cite{li2020widget}. This is a promising direction for future work.\\

\noindent \textbf{Beyond View Hierarchies.}
In our work, we exclusively derived our state representation from the view hierarchy and ignored visual information from the screen. While this was sufficient for solving the wide variety of problems in our benchmarks, we believe that considering the pixel information is an interesting avenue for future work. For example, in many cases, salient information is missing from the view hierarchy and can only be extracted from the screen's pixels. Previous work leveraged computer vision techniques (e.g., object detection and optical character recognition) to derive information from a smartphone screen, without relying on the underlying view hierarchy \cite{sereshkeh2020vasta}.

%% file: sections/conclusion.tex
\section{Concluding Remarks}

Building smarter smartphones has the potential to broaden accessibility of phone applications and improve the user experience. In this paper, we explored the use of RL to learn to accomplish tasks using mobile apps. RL agent states were derived from the underlying representation of on-screen elements, rewards were based on progress made in the task, and agents could interact with elements on the phone screen by tapping or typing. Our experiments showed that our RL agents could learn to accomplish multi-step tasks in a number of mobile apps, as well as achieve modest generalization across different mobile apps. An important contribution of our work was the development of a mobile app RL environment that is compatible with OpenAI Gym and its provision through the \our training platform.
The release of this training platform together with a suite of benchmarks opens the door to further research into learning to accomplish a diversity of tasks using mobile apps.

%% file: sections/supp.tex
{
\begin{center}
\huge{\our: Learning to Accomplish Tasks in \\Mobile Apps via Reinforcement Learning} \\
\vspace{1ex}
  \Large{Technical Appendix}
  \vspace{2ex}
\end{center}
}

\section{Experimental Setup}\label{app:exp_setup}

\textbf{Hardware Specification}\\
%
Our workstation, used both for training and testing, has the Intel(R) Xeon(R) CPU E5-2670 v2 @ 2.50GHz with 20 CPU cores. The total amount of memory is 503 GB. The workstation also has 8 NVIDIA Tesla M40 GPUs, each with 24 GB of GPU memory. The operating system on the workstation is CentOS 7.0.  

\textbf{Hyperparameters}\\
The following hyperparameters are used in all experiments presented in this paper. Hyperparameters not listed here were assigned the default values used in \cite{baselines}. For the policy network, we used a fully-connected MLP with \textbf{3} hidden layers of \textbf{1024} units, and tanh nonlinearities. $n$, the fixed number of on-screen elements for our state representation, was set empirically to 20. 







\begin{table}[h!]
\centering
\resizebox{0.4\textwidth}{!}{%
\begin{tabular}{|l|c|}
\hline
\textbf{Hyperparameter}          & \textbf{Value} \\ \hline
Number of epochs           & 4            \\ \hline
Learning rate                    & $3\times10^{-4}$            \\ \hline
Minibatch size                      & \# of emulators (3 or 35)            \\ \hline
Discount ($\gamma$)                & 0.99            \\ \hline
VF coefficient & 0.5            \\ \hline
Clipping range      & 0.2            \\ \hline
Entropy coefficient   & 0.01            \\ \hline
\end{tabular}%
}
\caption{Hyperparameters for PPO2 used in our experiments.}
\label{tab:my-table}
\end{table}

\textbf{Intermediate Rewards used in the Tasks}\\

The following describes the intermediate rewards given to the agent in each task.

\begin{itemize}
    \item Settings - Easy
    \begin{itemize}
        \item No intermediate rewards
    \end{itemize}
    \item Settings - Medium
    \begin{itemize}
        \item The agent is rewarded after navigating to Wi-Fi settings screen
    \end{itemize}
    \item Settings - Hard
    \begin{itemize}
        \item The agent is rewarded after navigating to Wi-Fi settings screen and after navigating to add new Wi-Fi network screen
    \end{itemize}
    \item Split - Easy
    \begin{itemize}
        \item The agent is rewarded after navigating to the `create new group' screen and after entering the correct token into the `new group name' editable field
    \end{itemize}
    \item Split - Medium
    \begin{itemize}
        \item The agent is rewarded with the intermediate rewards of the easy split task. The agent is also rewarded when it successfully creates a new group with the correct name, after navigating to the group screen, after navigating to the `add new member' screen, and after entering the correct token into the `new member name' editable field
    \end{itemize}
    \item Split - Hard
    \begin{itemize}
        \item The agent is rewarded with the intermediate rewards of the easy and medium split tasks. The agent is also rewarded when it successfully adds a member to the group with the correct name, after navigating to the `add new expense' screen, after entering either the correct amount of money or the correct expense name, and after entering both the correct amount of money or the correct expense name (in this case, the agent gets a slightly larger reward)
    \end{itemize}
    \item Alarm - Easy
    \begin{itemize}
        \item No intermediate rewards
    \end{itemize}
    \item Alarm - Medium
    \begin{itemize}
        \item The agent is rewarded when first alarm clock is set properly
    \end{itemize}
    \item Alarm - Hard
    \begin{itemize}
        \item The agent is rewarded when first alarm clock is set, when second alarm is set, when third alarm is set, and is rewarded a larger intermediate reward when any two alarm clocks are set
    \end{itemize}
    \item Shopping - Easy
    \begin{itemize}
        \item No intermediate rewards
    \end{itemize}
    \item Shopping - Medium
    \begin{itemize}
        \item The agent is rewarded when navigating to `more options' screen, and when navigating to `add new list' screen
    \end{itemize}
    \item Shopping - Hard
    \begin{itemize}
        \item The agent is rewarded when navigating to `more options' screen, when navigating to `add new list' screen, and when successfully adding a new list with the correct name
    \end{itemize}
\end{itemize}

\section{Additional Results}\label{app:more_results}

Here we include additional results from our experiments.

\subsection{Multi-step task learning}

\begin{figure*}
    \centering
    \subfigure[]{
		\includegraphics[width=0.3\textwidth]{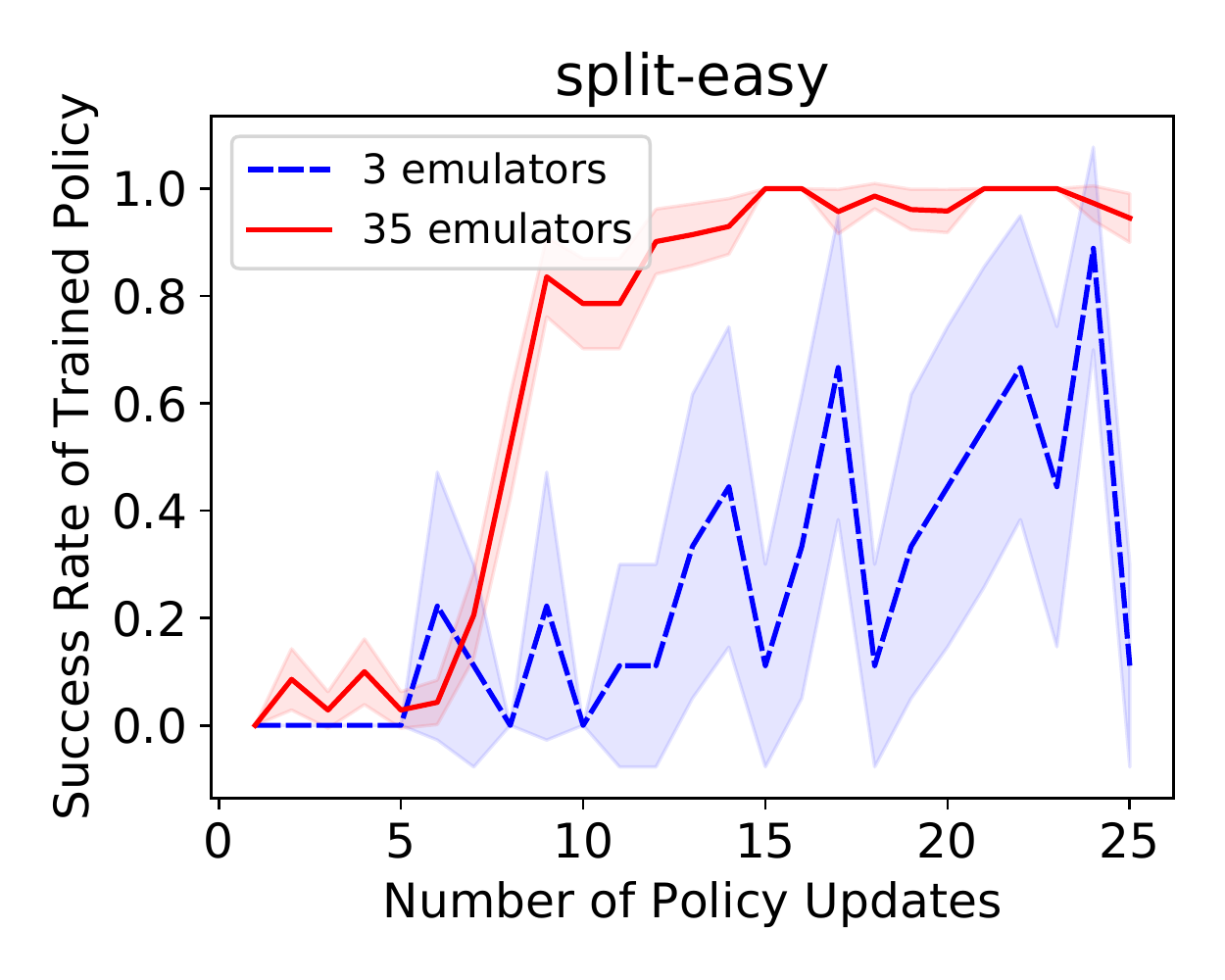}
		\label{fig:split-easy_3_35}
	}
 	\hspace{2pt}
	\subfigure[]{
		\includegraphics[width=0.3\textwidth]{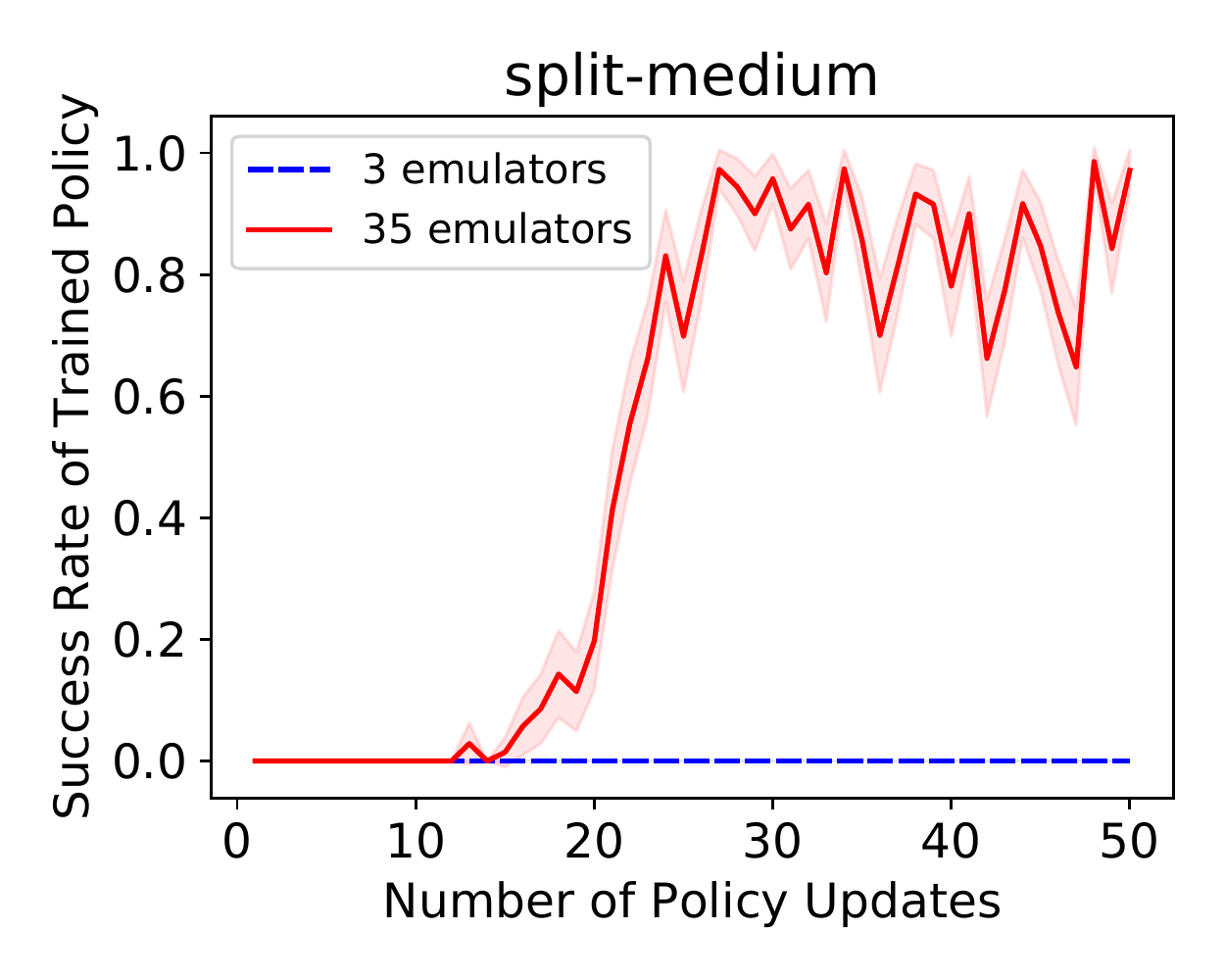}
		\label{fig:split-medium_3_35}
	}
	\hspace{2pt}
	\subfigure[]{
		\includegraphics[width=0.3\textwidth]{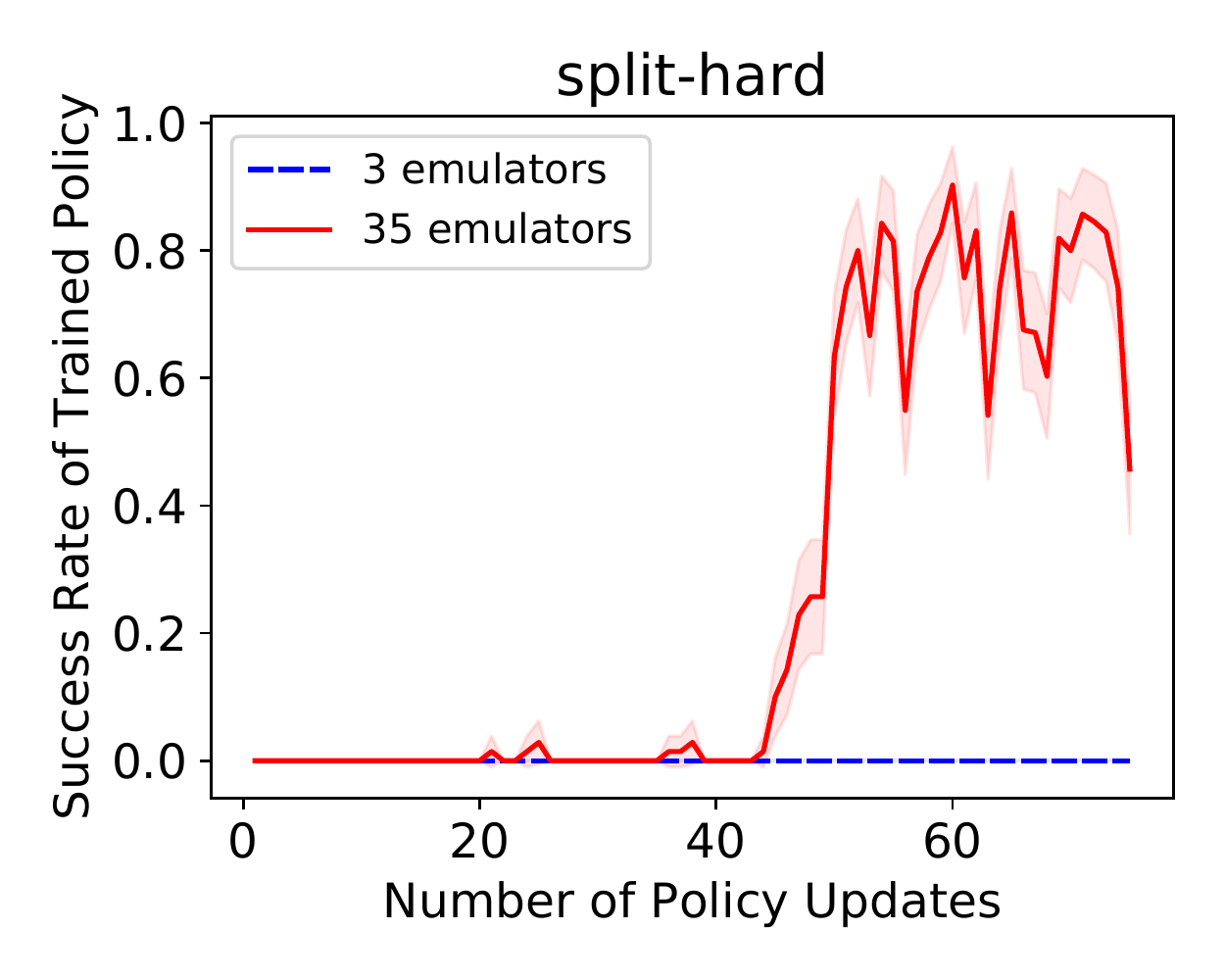}
		\label{fig:split-hard_3_35}
	}
	
	\subfigure[]{
		\includegraphics[width=0.3\textwidth]{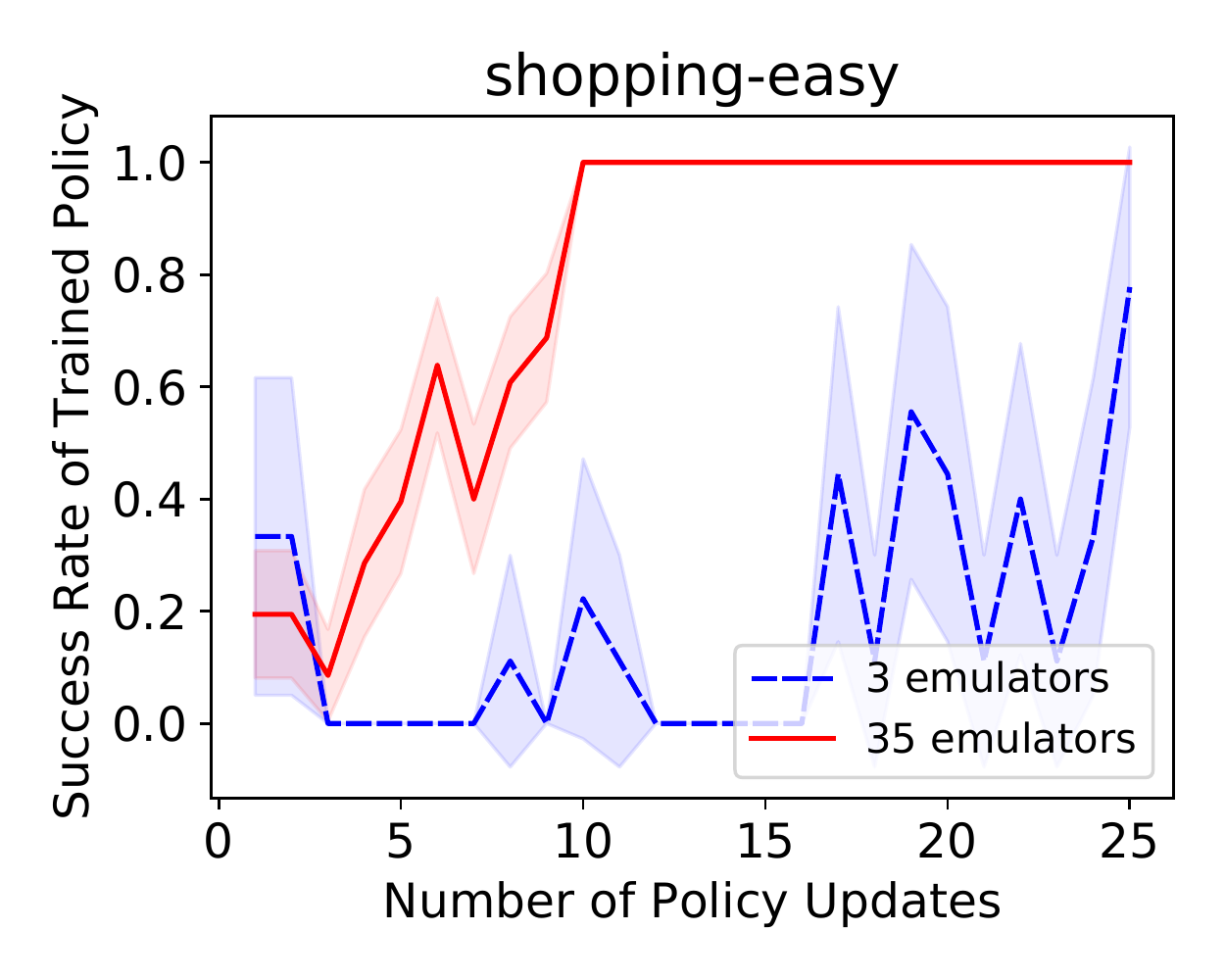}
	}
 	\hspace{2pt}
	\subfigure[]{
		\includegraphics[width=0.3\textwidth]{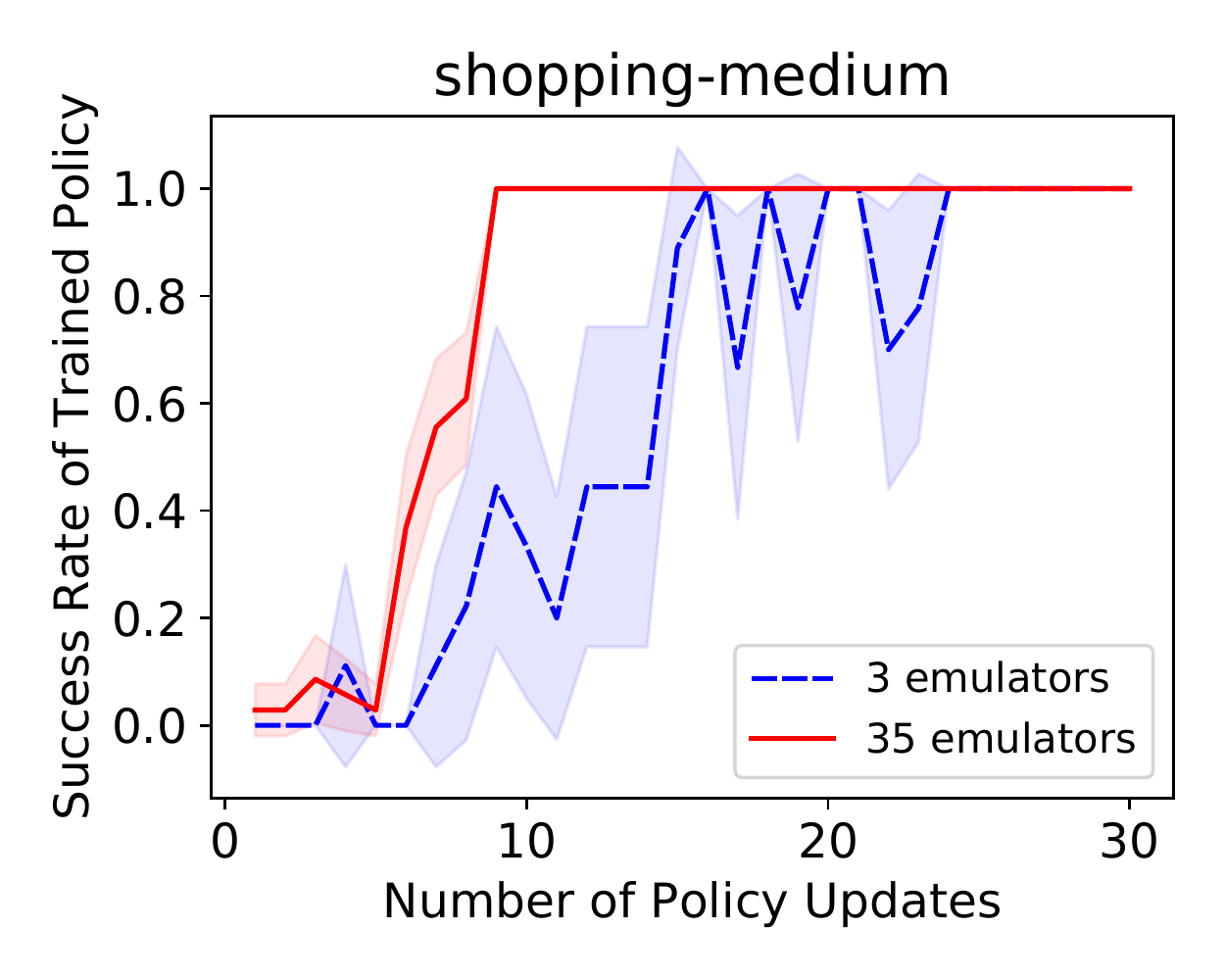}
	}
	\hspace{2pt}
	\subfigure[]{
		\includegraphics[width=0.3\textwidth]{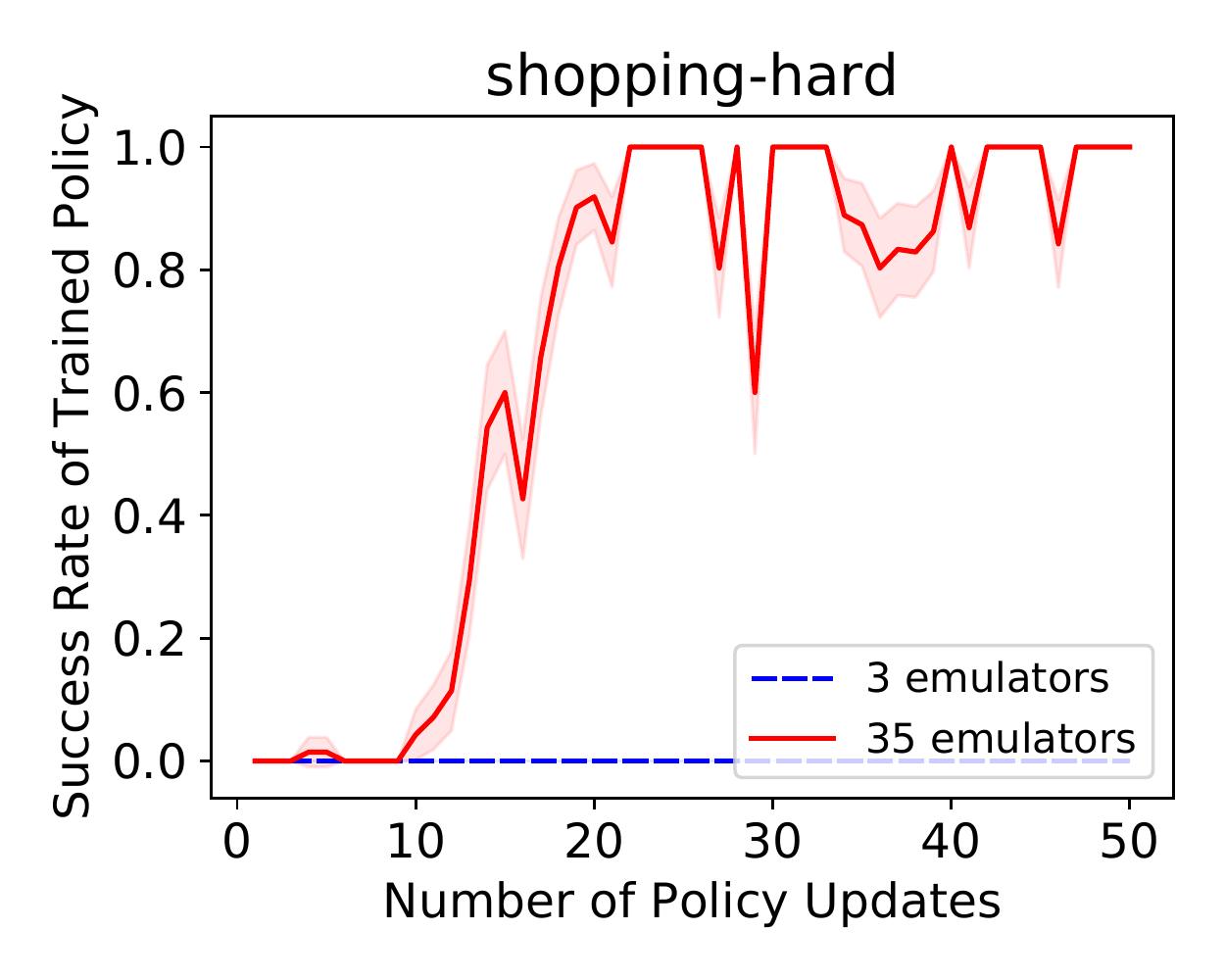}
		\label{fig:shopping-hard_3_35}
	}
	
	\subfigure[]{
		\includegraphics[width=0.3\textwidth]{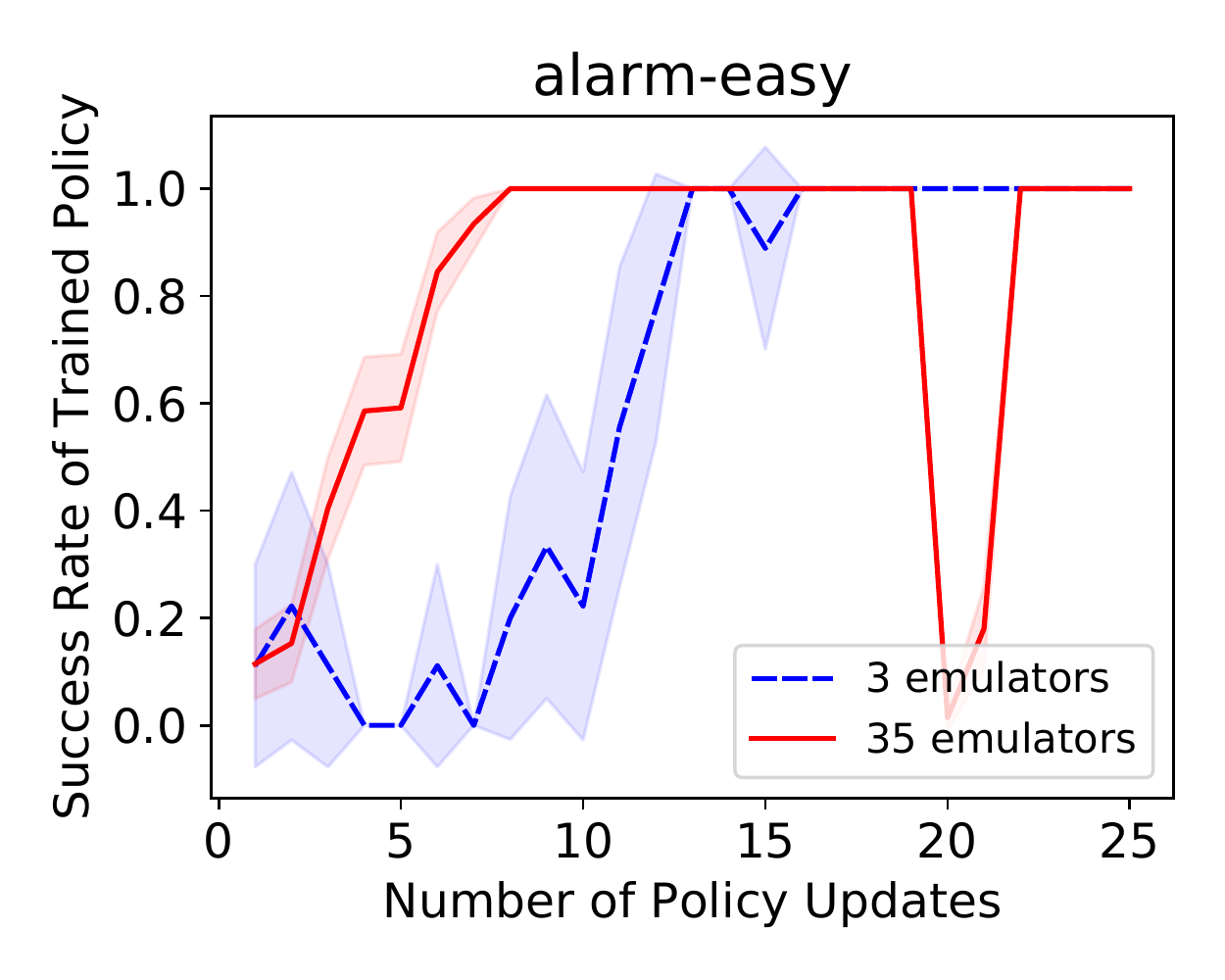}
	}
 	\hspace{2pt}
	\subfigure[]{
		\includegraphics[width=0.3\textwidth]{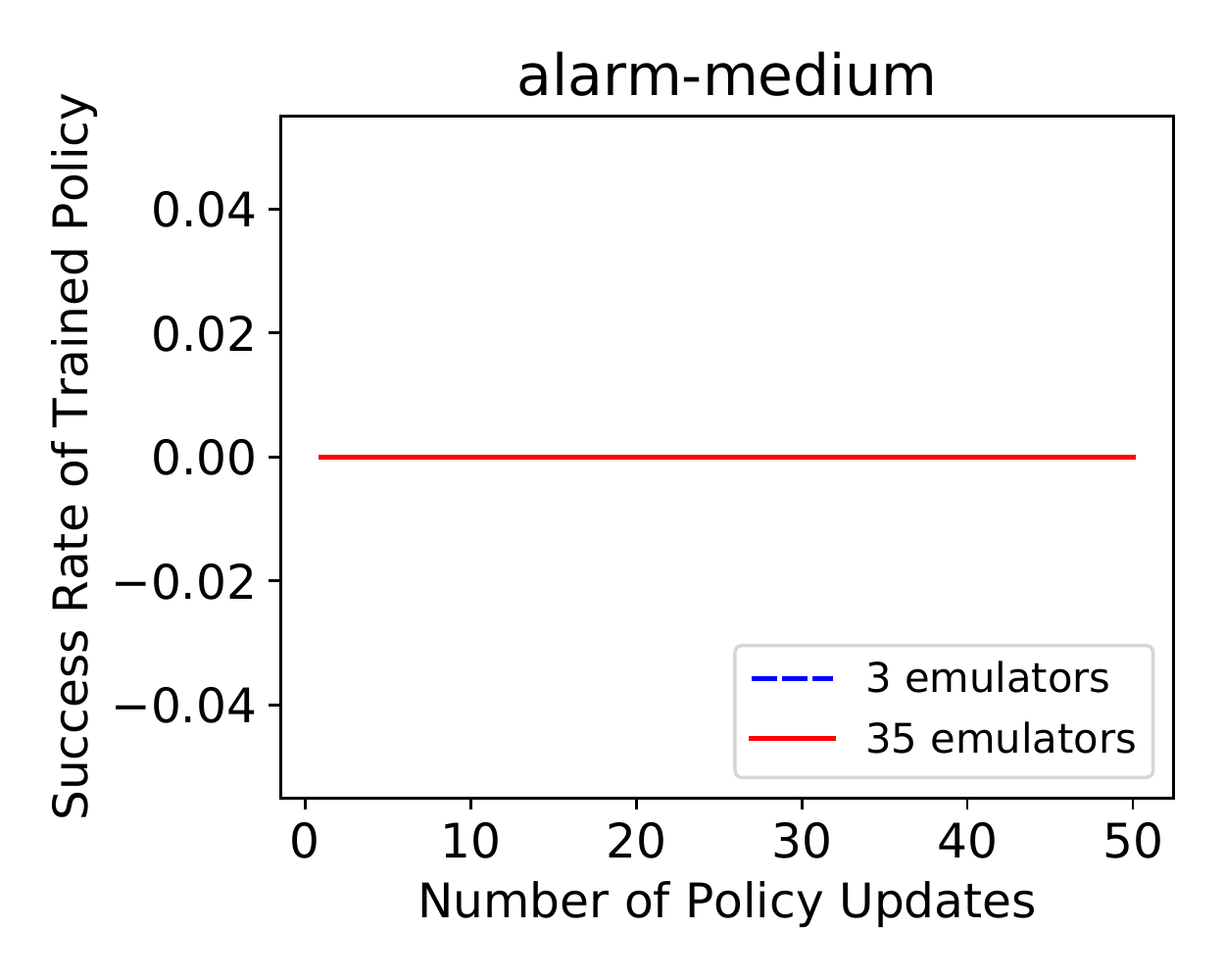}
	}
	\hspace{2pt}
	\subfigure[]{
		\includegraphics[width=0.3\textwidth]{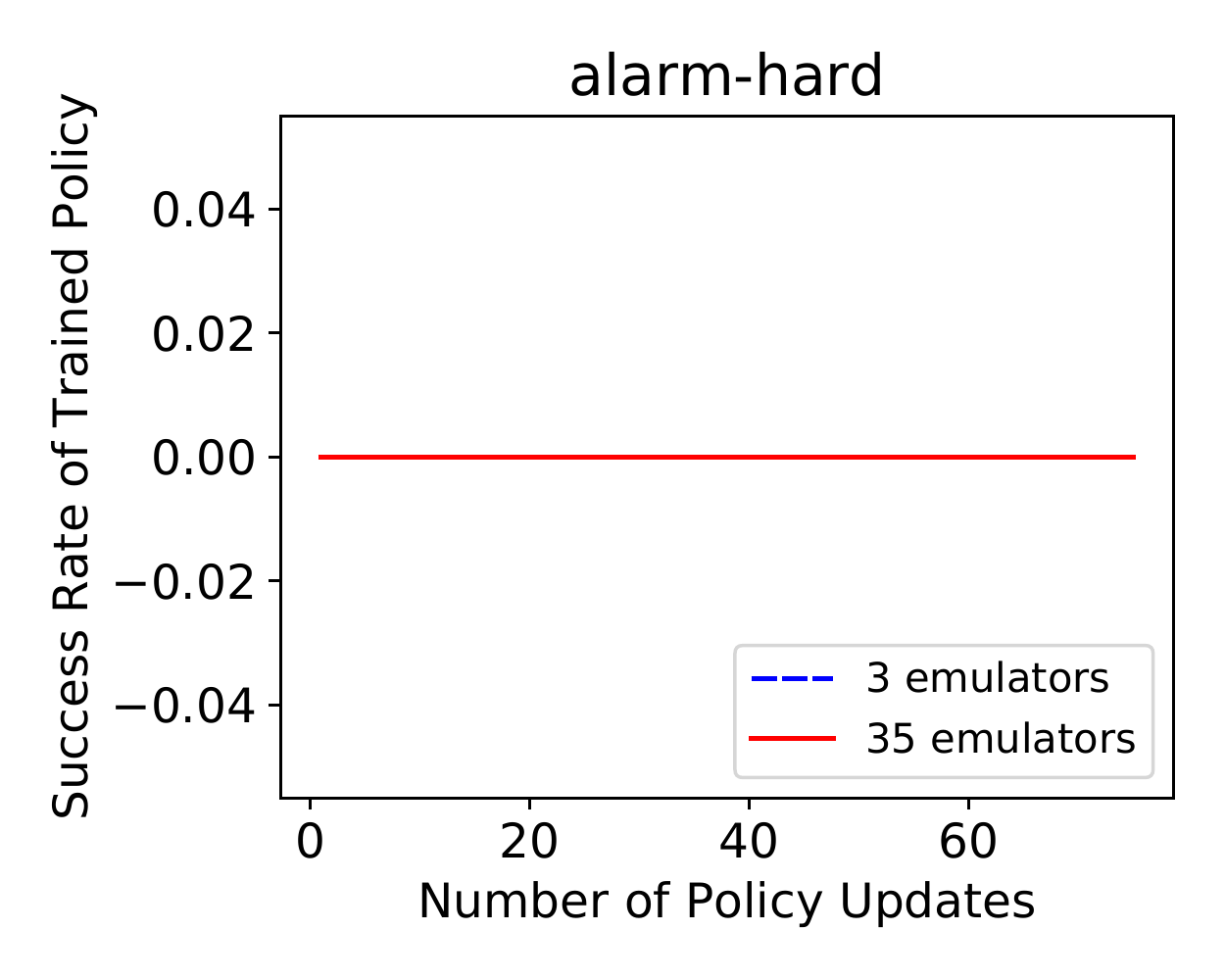}
	}
	
	\caption{The success rate of a trained policy (percentage of successful test episodes) as a function of the number of policy updates in various tasks. \textbf{Comparison between 3 and 35 emulators.}}
	\label{fig:3_35_extra}
	\squeezeup
\end{figure*}

\begin{figure*}
    \centering
    \subfigure[]{
		\includegraphics[width=0.3\textwidth]{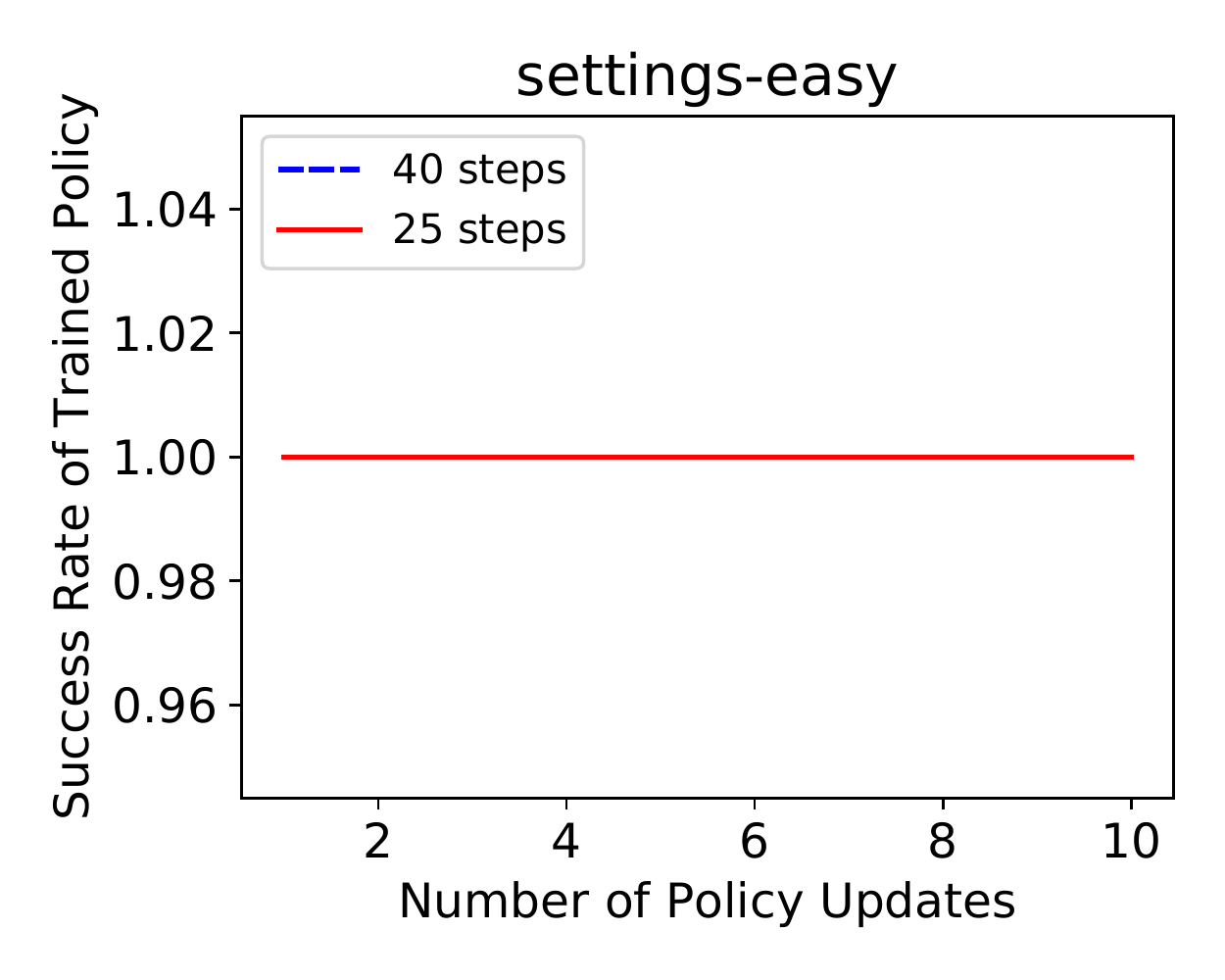}
	}
 	\hspace{2pt}
	\subfigure[]{
		\includegraphics[width=0.3\textwidth]{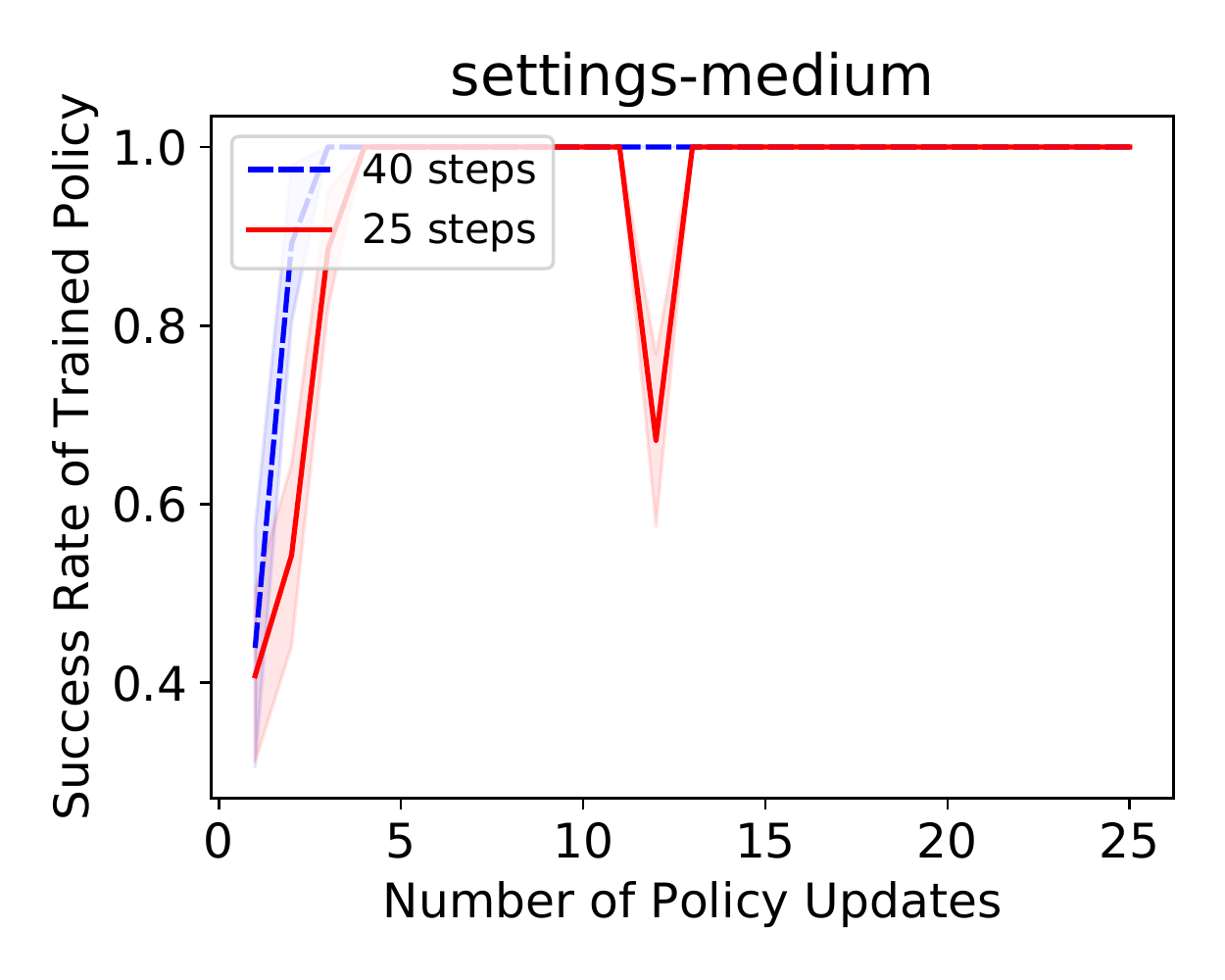}
	}
	\hspace{2pt}
	\subfigure[]{
		\includegraphics[width=0.3\textwidth]{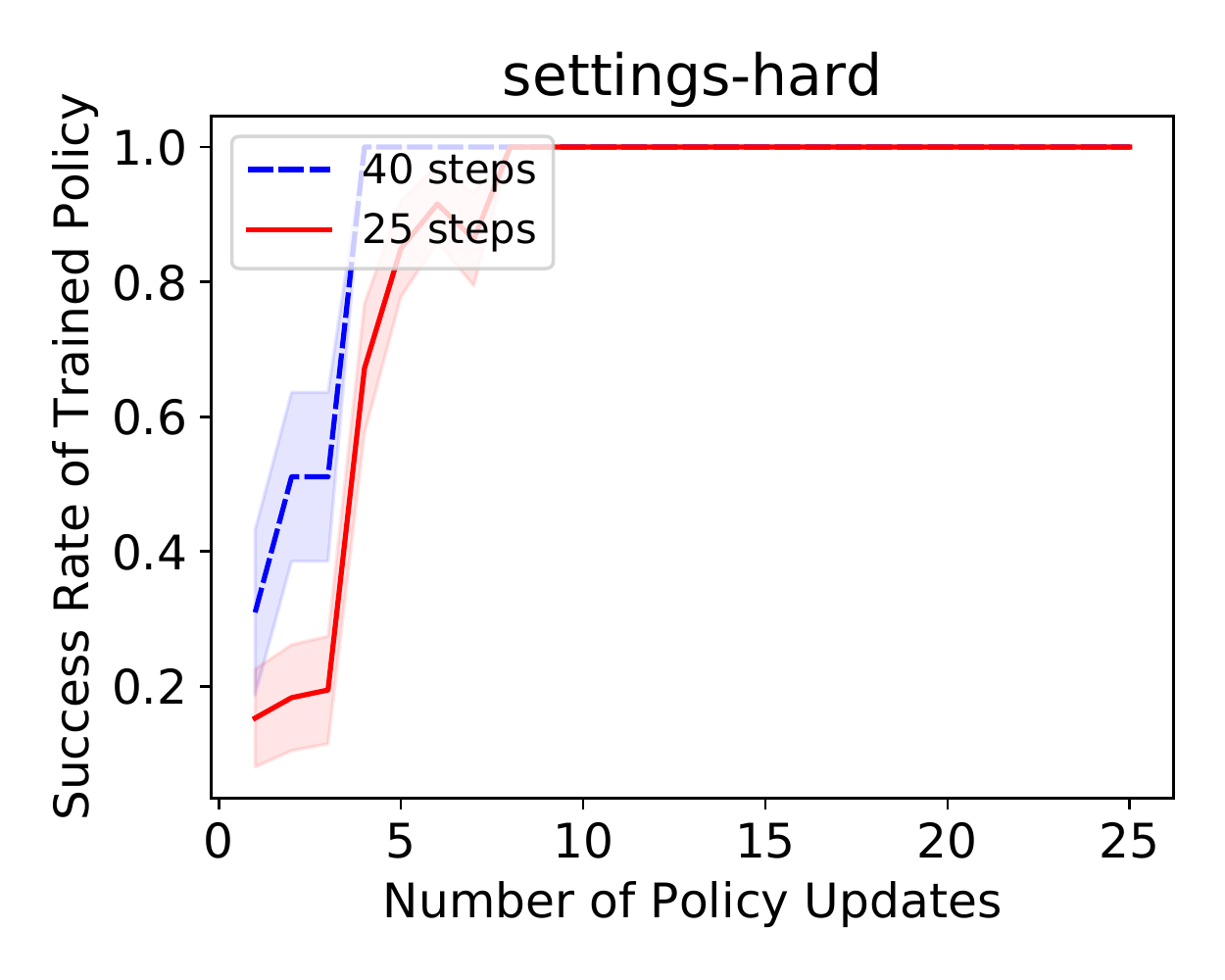}
	}
	
	\subfigure[]{
		\includegraphics[width=0.3\textwidth]{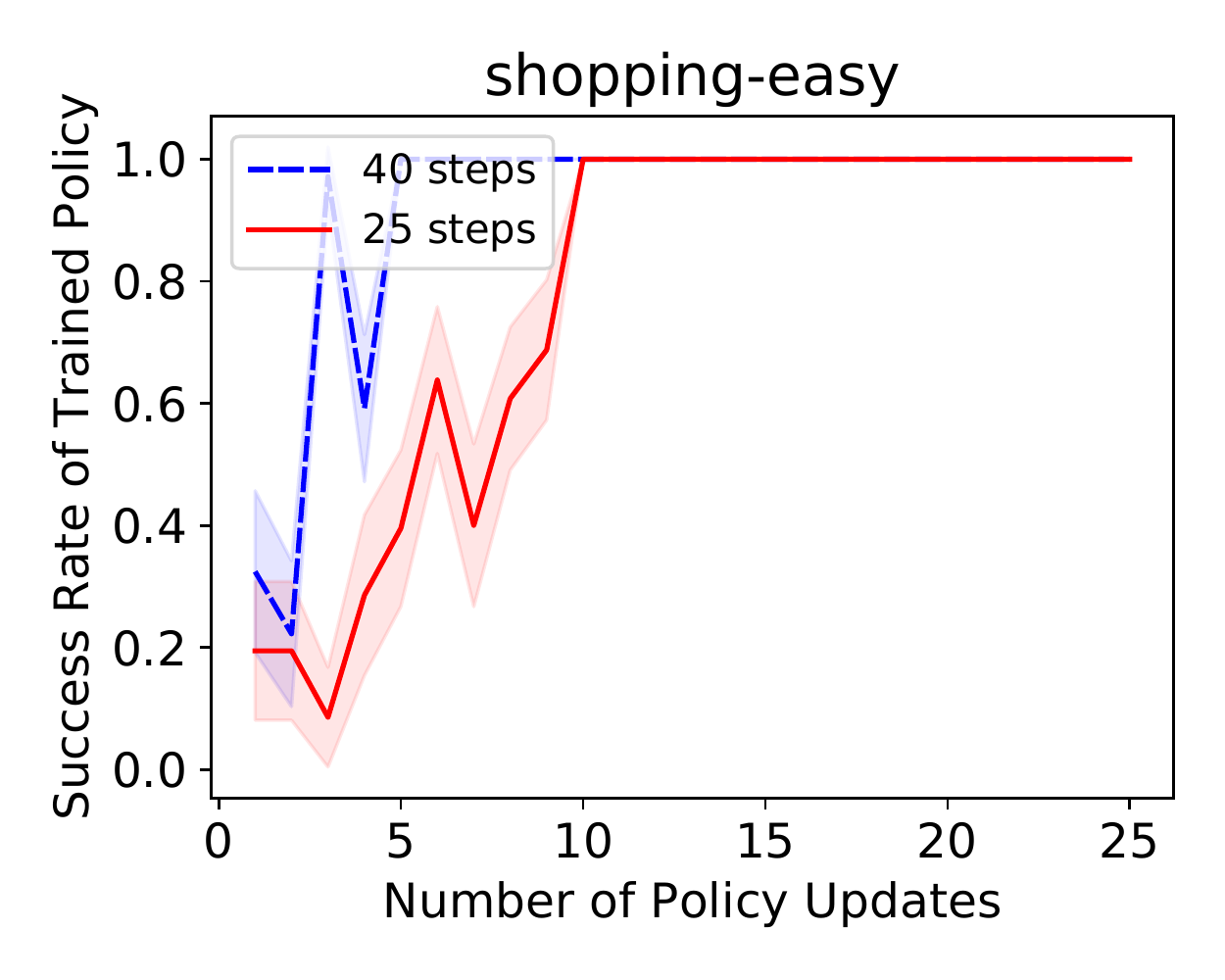}
	}
 	\hspace{2pt}
	\subfigure[]{
		\includegraphics[width=0.3\textwidth]{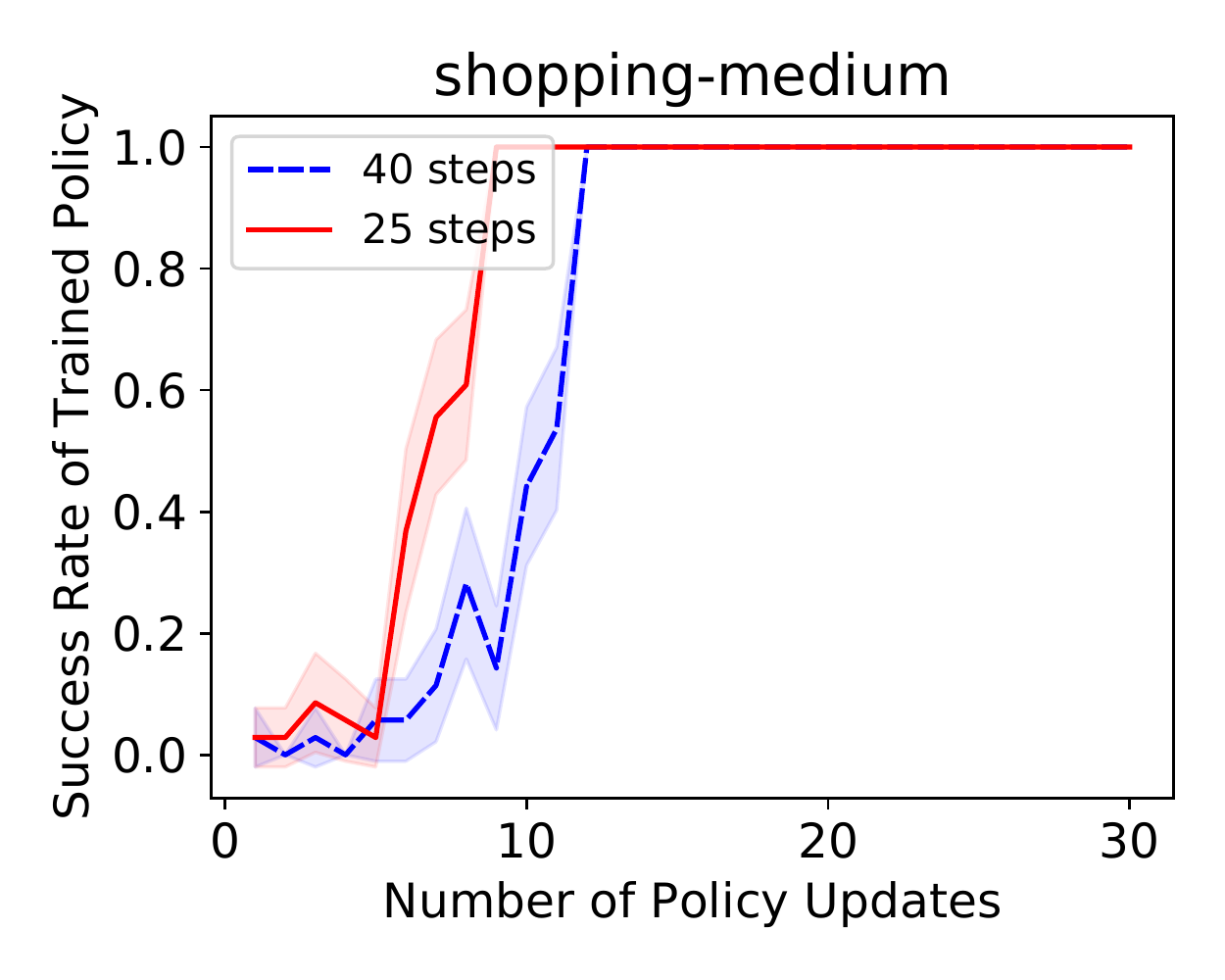}
	}
	\hspace{2pt}
	\subfigure[]{
		\includegraphics[width=0.3\textwidth]{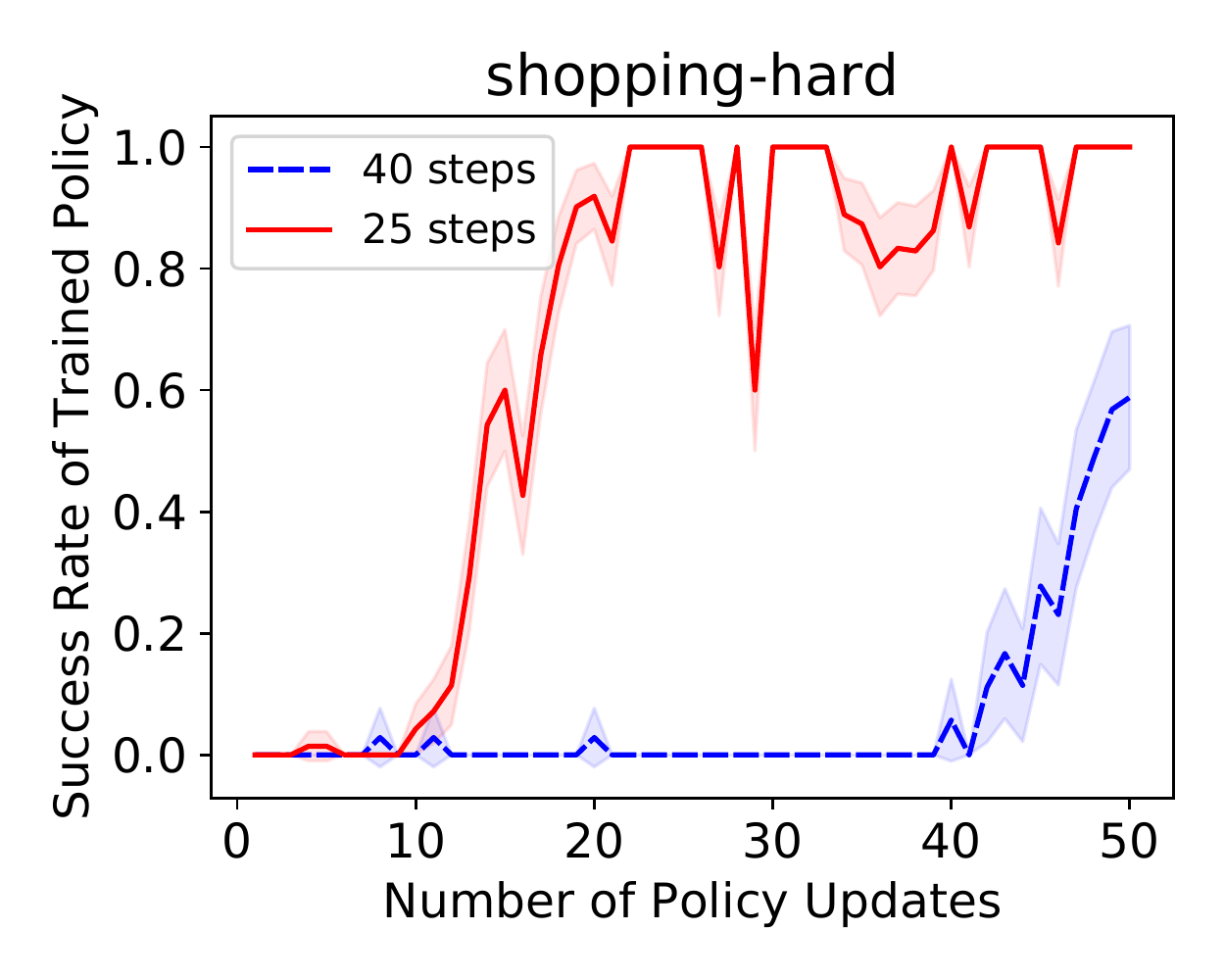}
	}
	
	\subfigure[]{
		\includegraphics[width=0.3\textwidth]{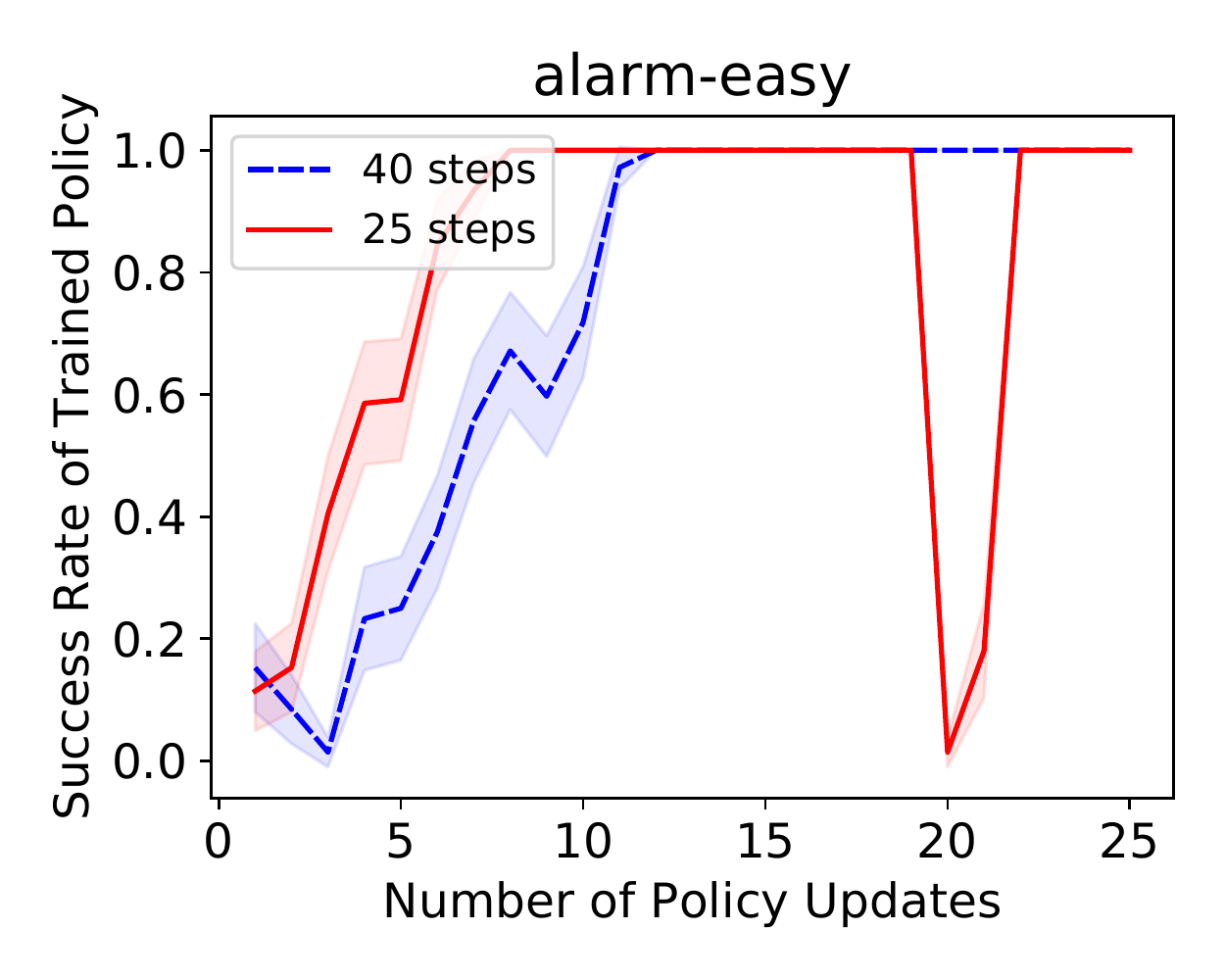}
	}
 	\hspace{2pt}
	\subfigure[]{
		\includegraphics[width=0.3\textwidth]{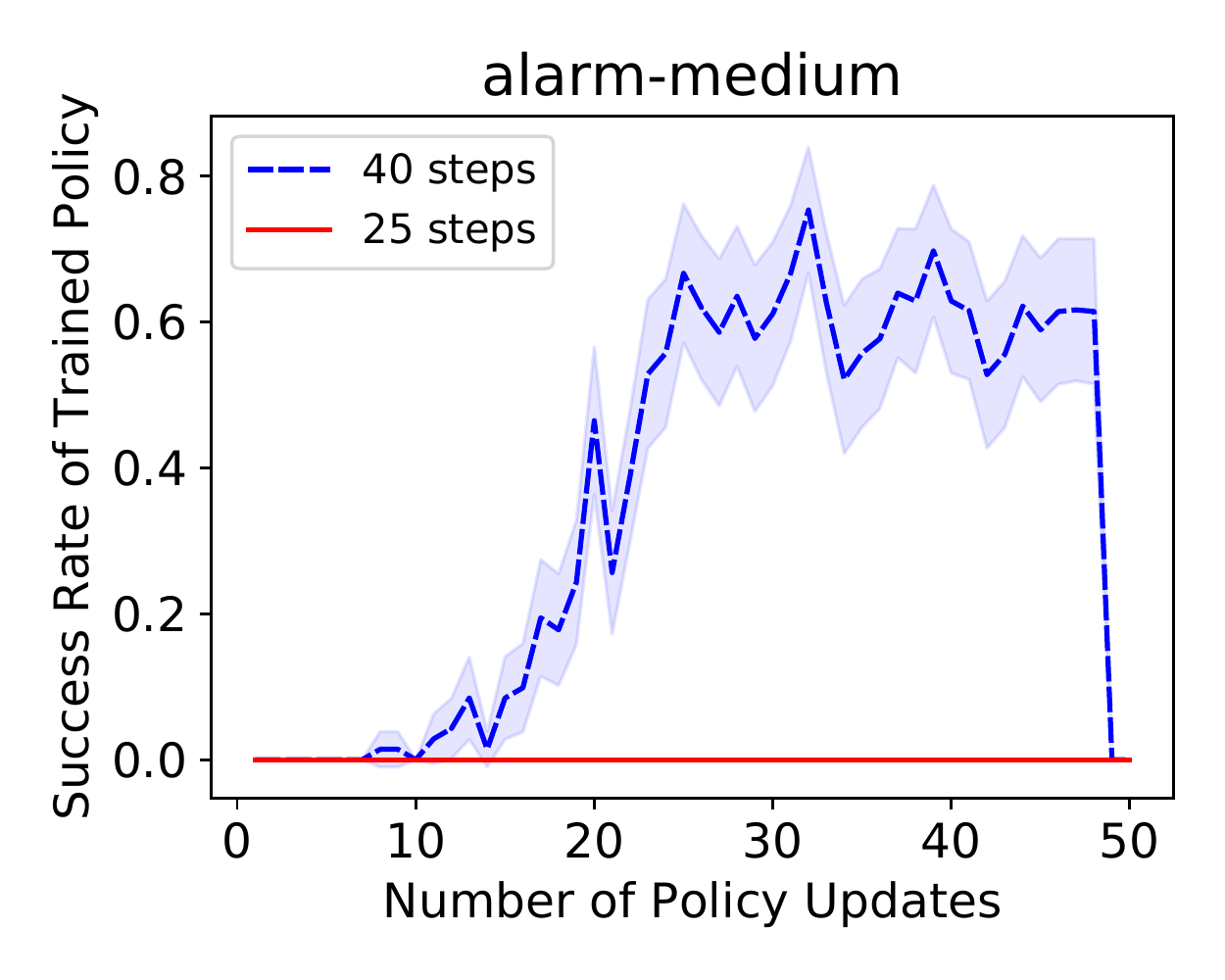}
		\label{fig:alarm_medium_25_40}
	}
	\hspace{2pt}
	\subfigure[]{
		\includegraphics[width=0.3\textwidth]{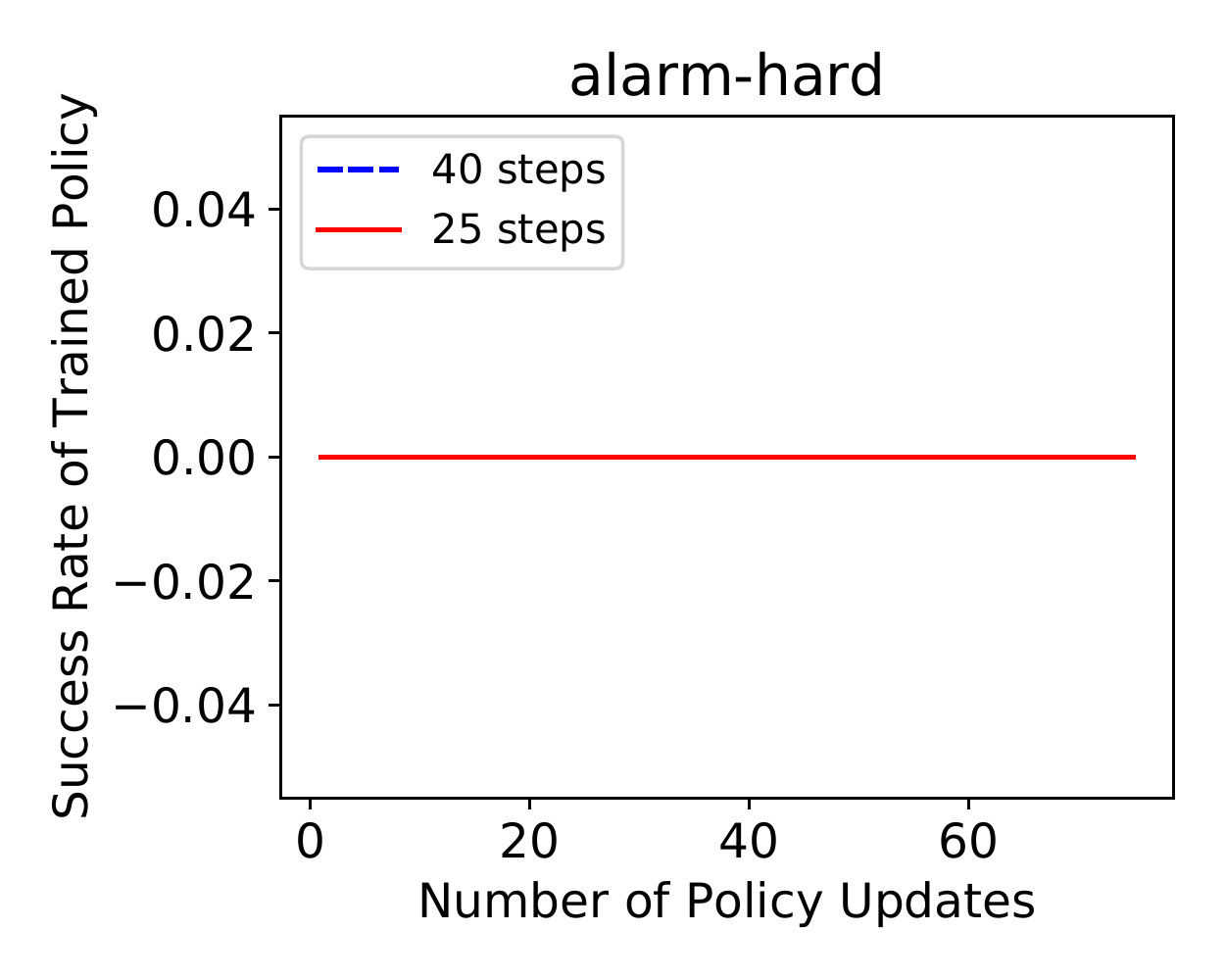}
	}
	
	\caption{The success rate of a trained policy (percentage of successful test episodes) as a function of the number of policy updates in various tasks. \textbf{Comparison between 25 and 40 steps per episode.}}
	\label{fig:25_40_extra}
	\squeezeup
\end{figure*}

\begin{figure*}
    \centering
    	\subfigure[]{
		\includegraphics[width=0.3\textwidth]{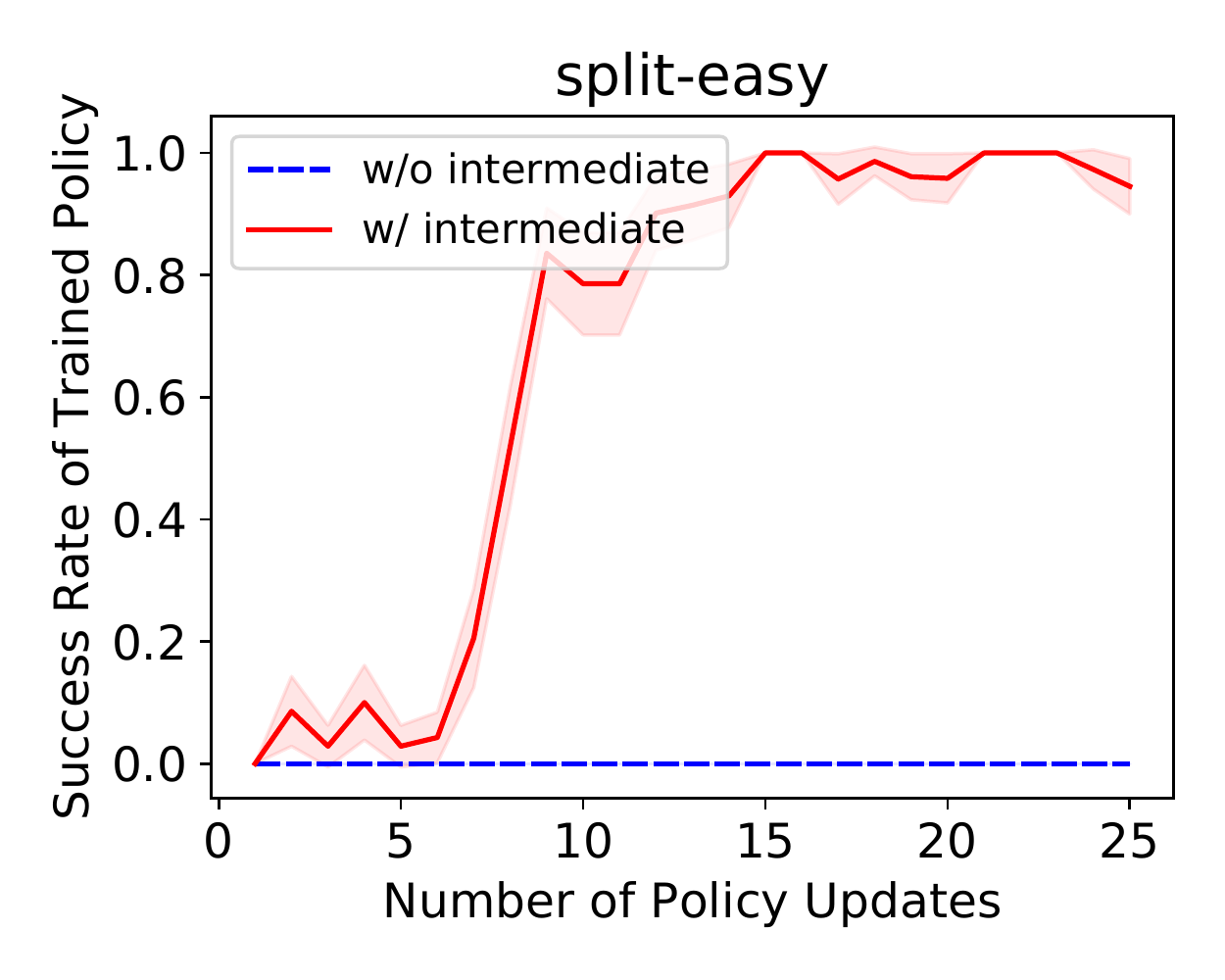}
	}
 	\hspace{2pt}
	\subfigure[]{
		\includegraphics[width=0.3\textwidth]{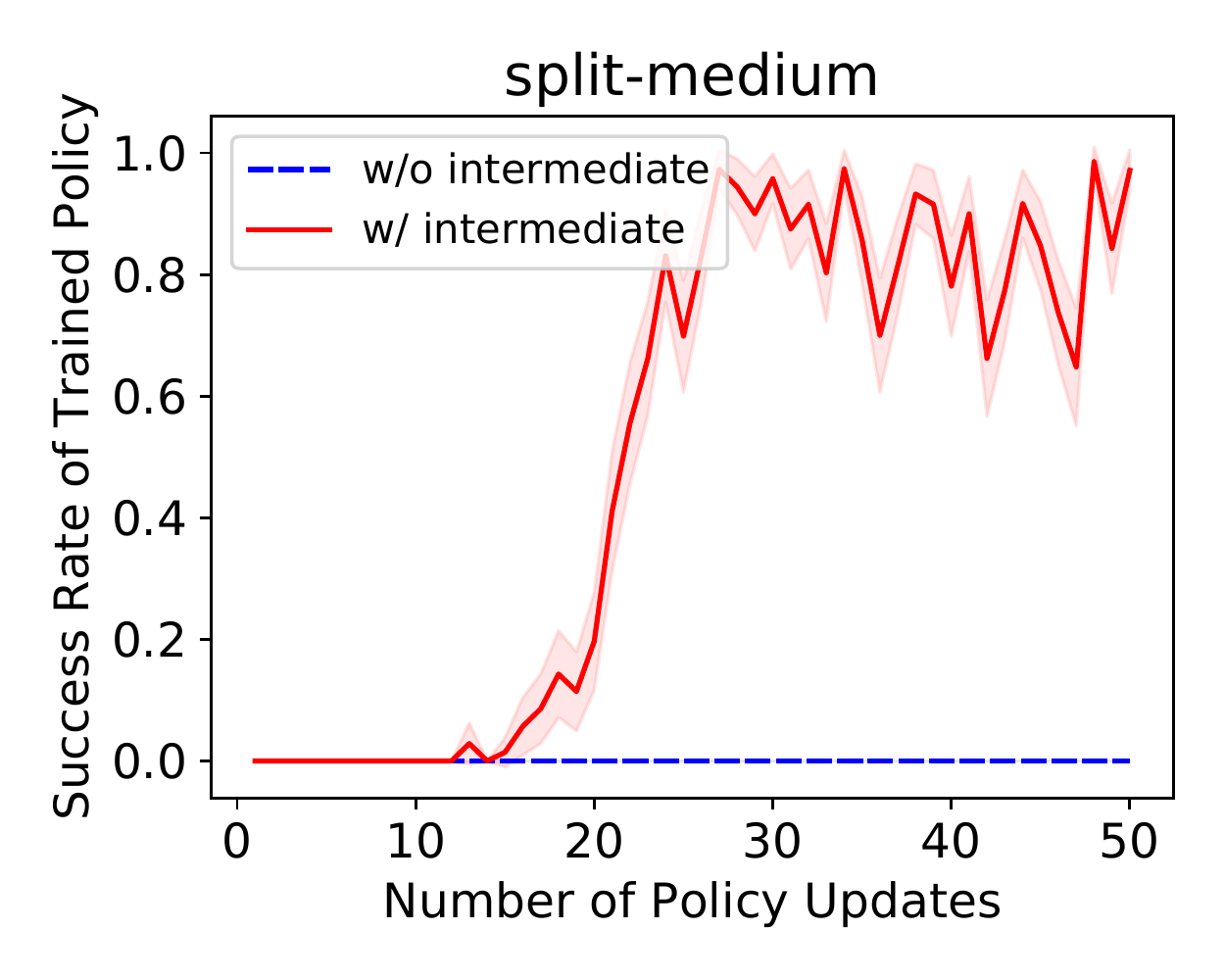}
	}
	\hspace{2pt}
	\subfigure[]{
		\includegraphics[width=0.3\textwidth]{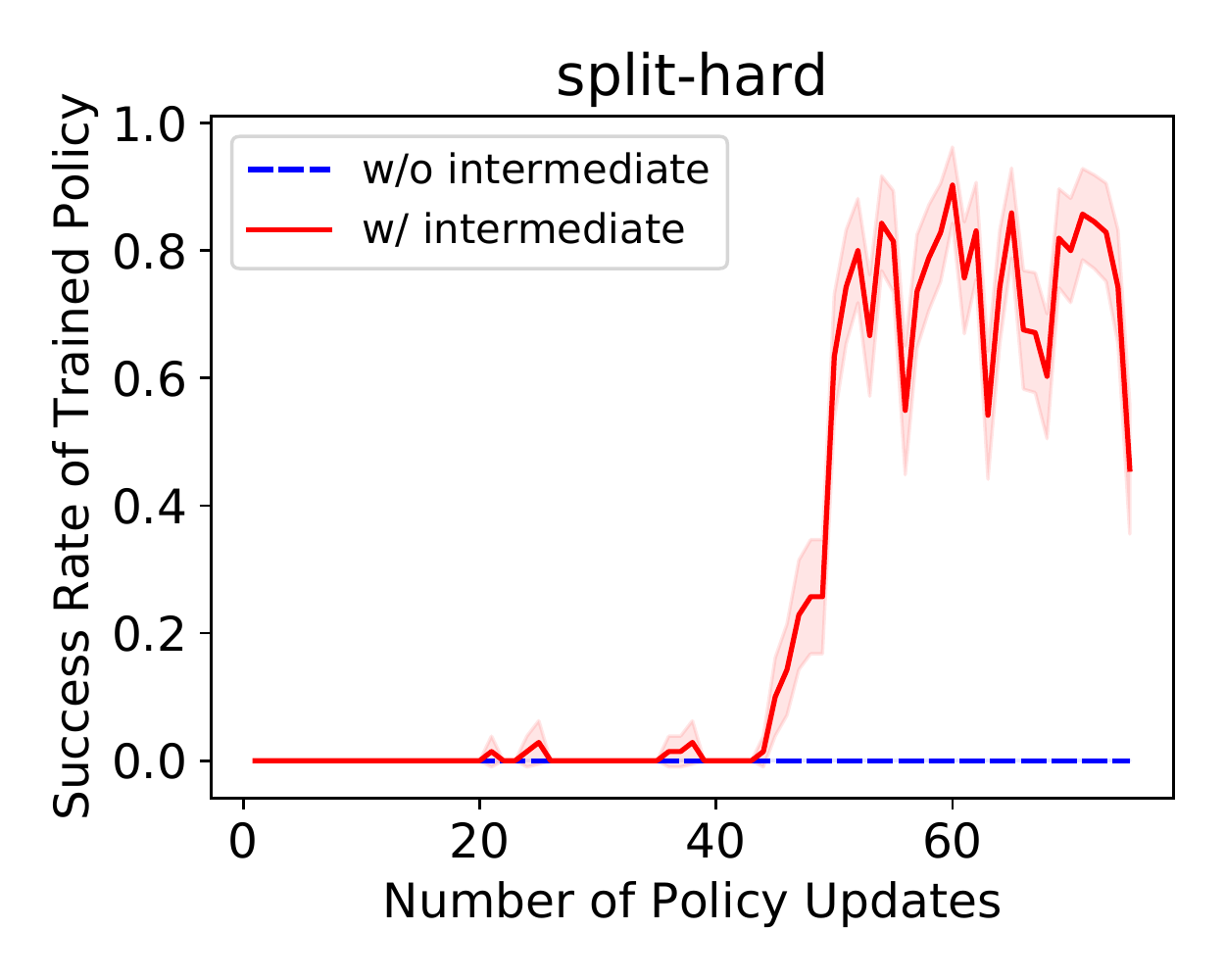}
	}
	
	\subfigure[]{
		\includegraphics[width=0.3\textwidth]{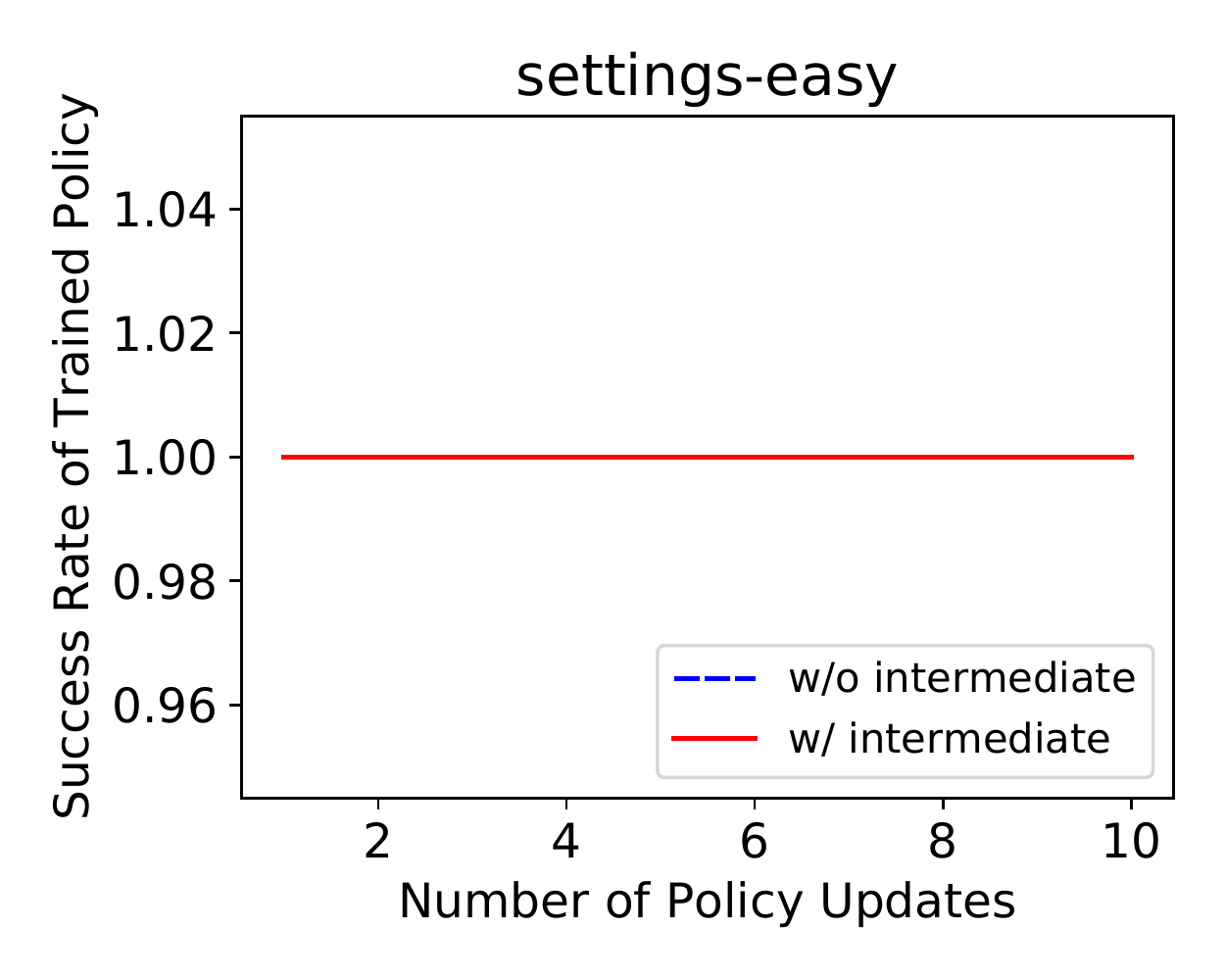}
	}
 	\hspace{2pt}
	\subfigure[]{
		\includegraphics[width=0.3\textwidth]{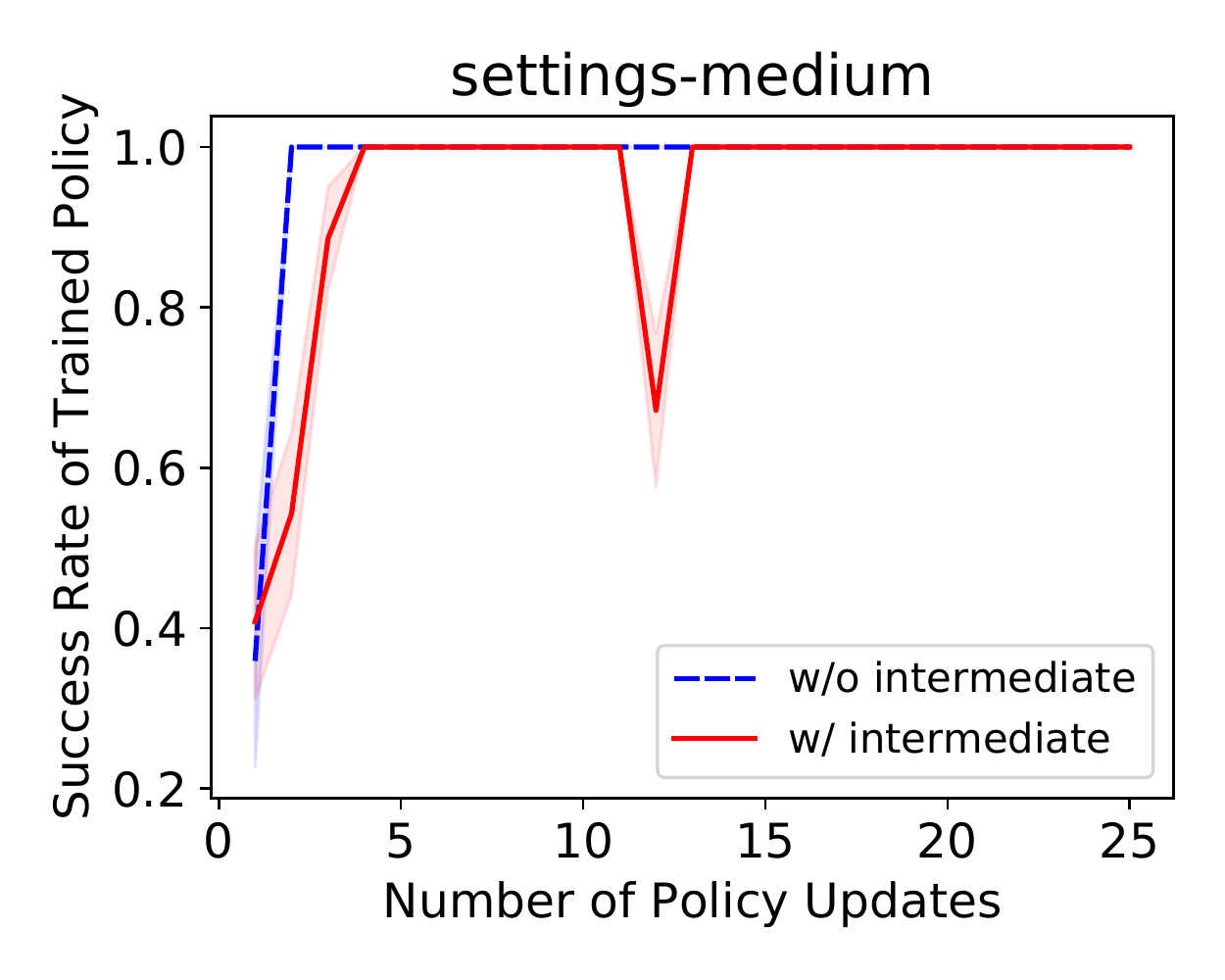}
	}
	\hspace{2pt}
	\subfigure[]{
		\includegraphics[width=0.3\textwidth]{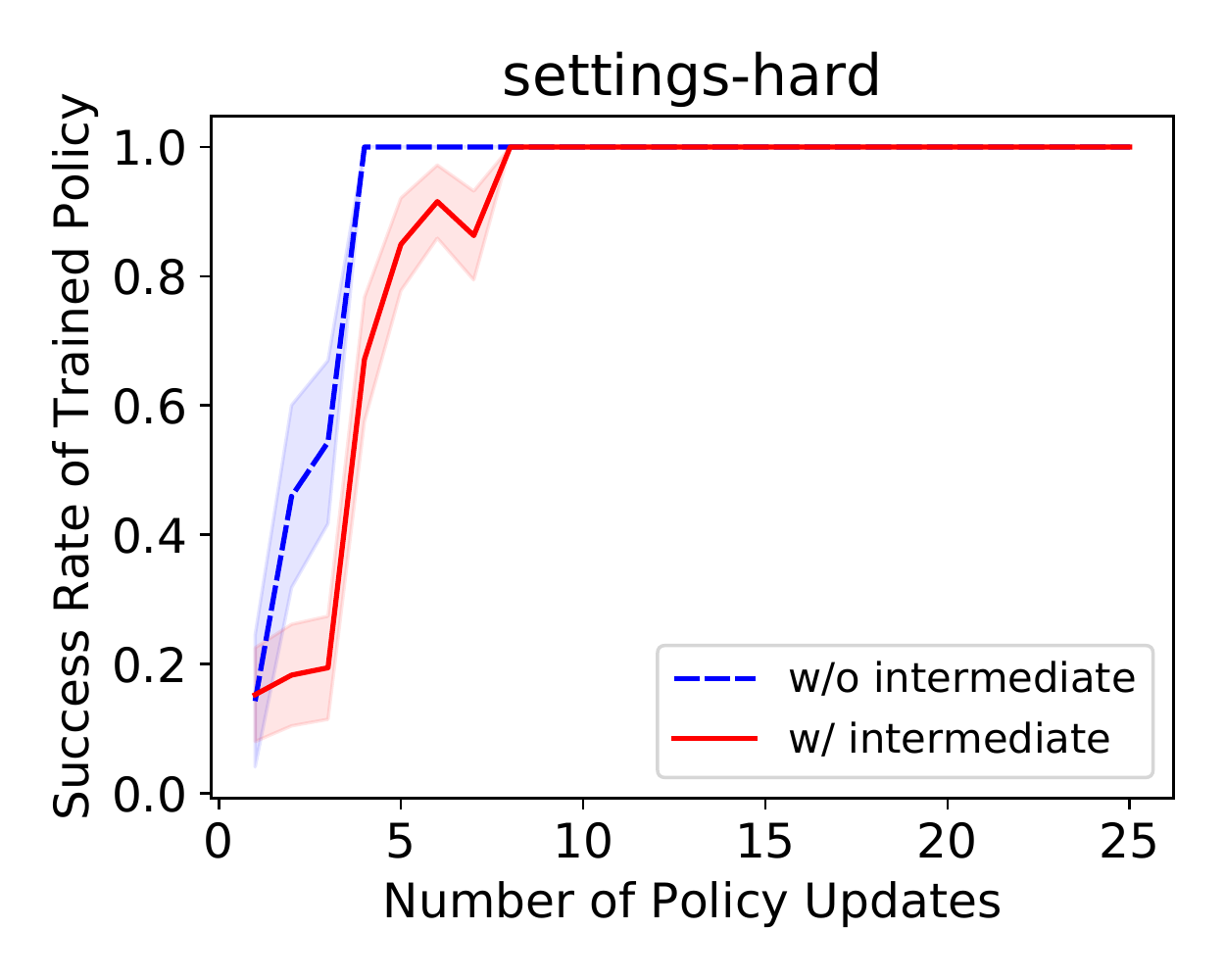}
		\label{fig:alarm-hard_3_35}
	}
	
	\subfigure[]{
		\includegraphics[width=0.3\textwidth]{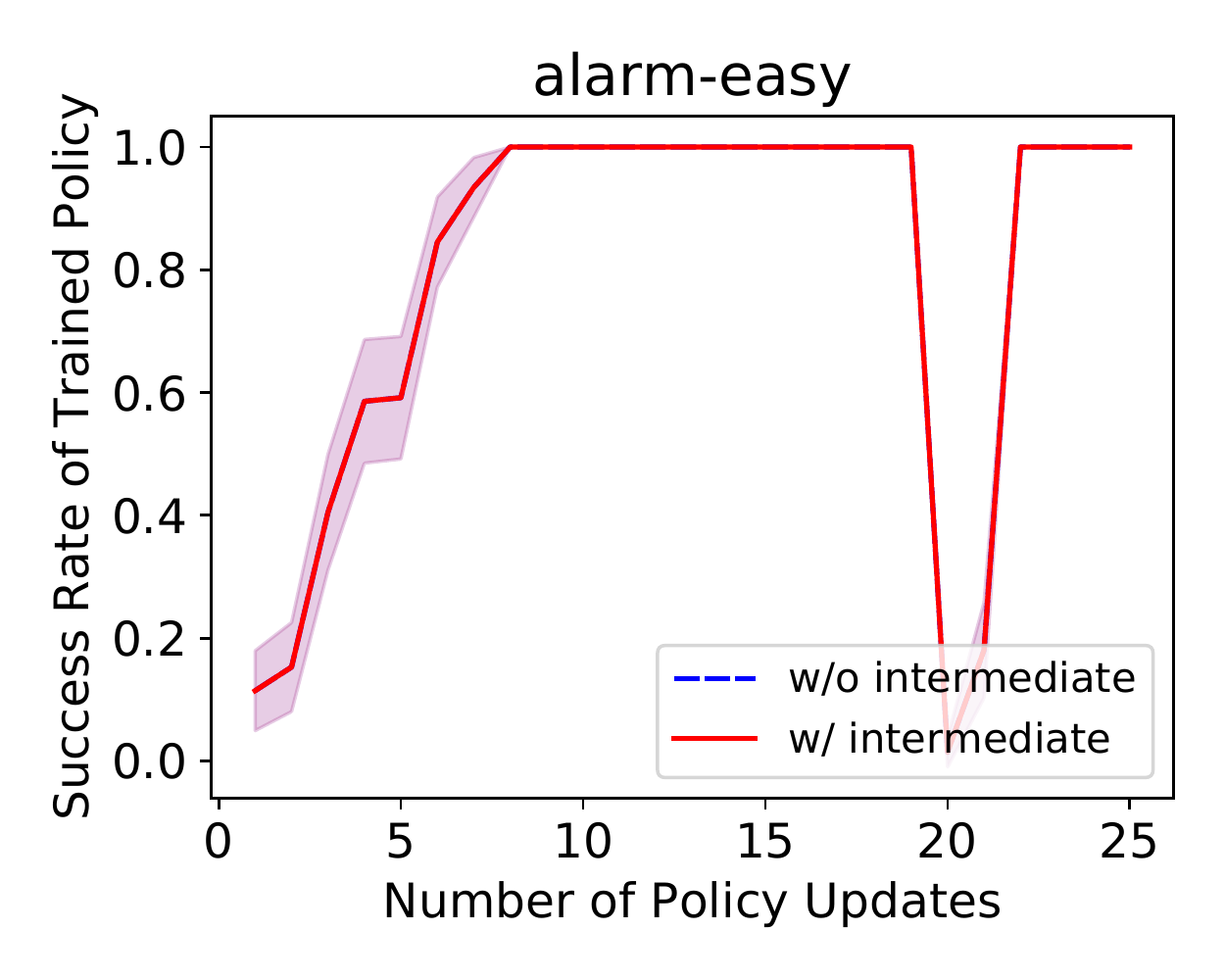}
	}
 	\hspace{2pt}
	\subfigure[]{
		\includegraphics[width=0.3\textwidth]{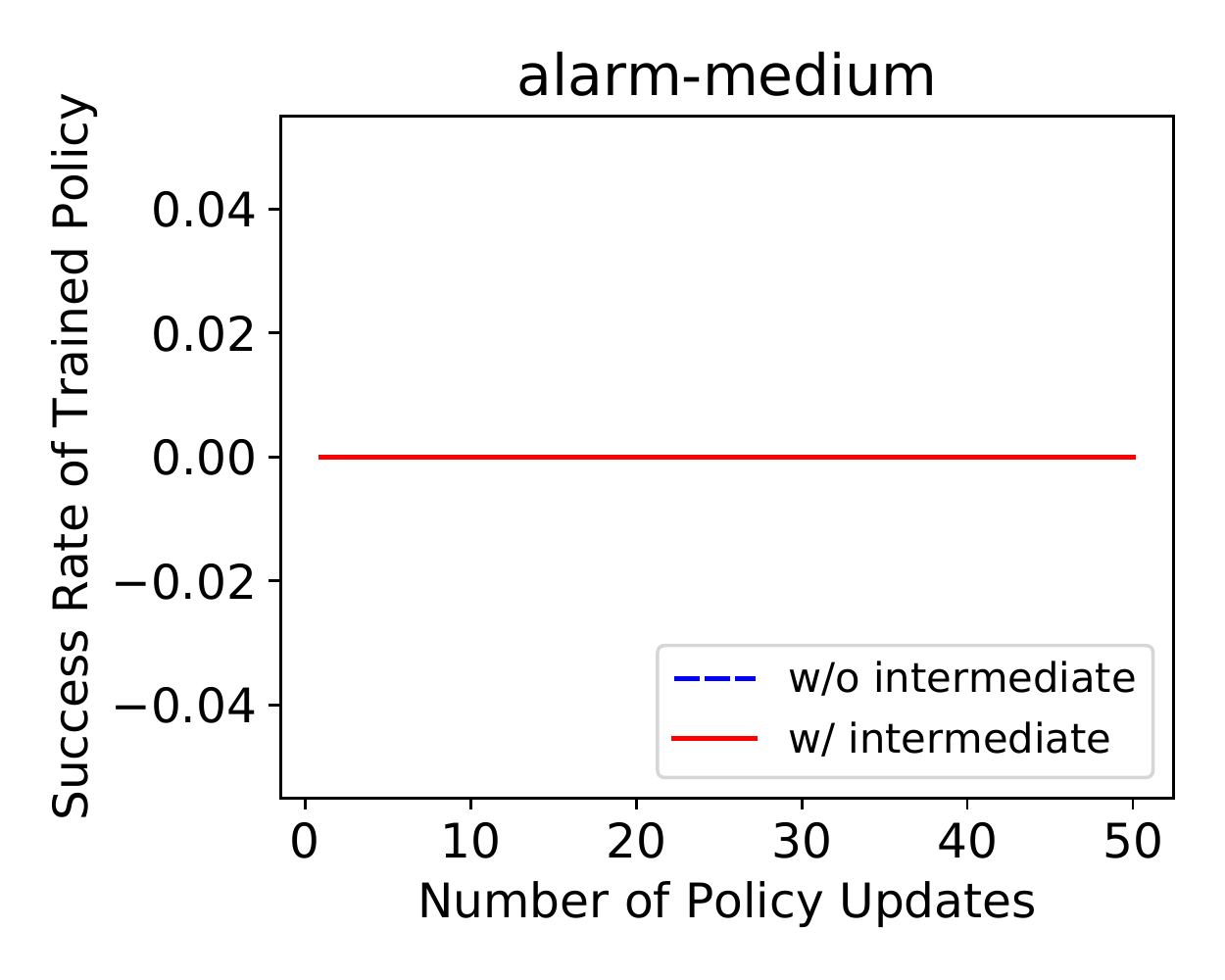}
	}
	\hspace{2pt}
	\subfigure[]{
		\includegraphics[width=0.3\textwidth]{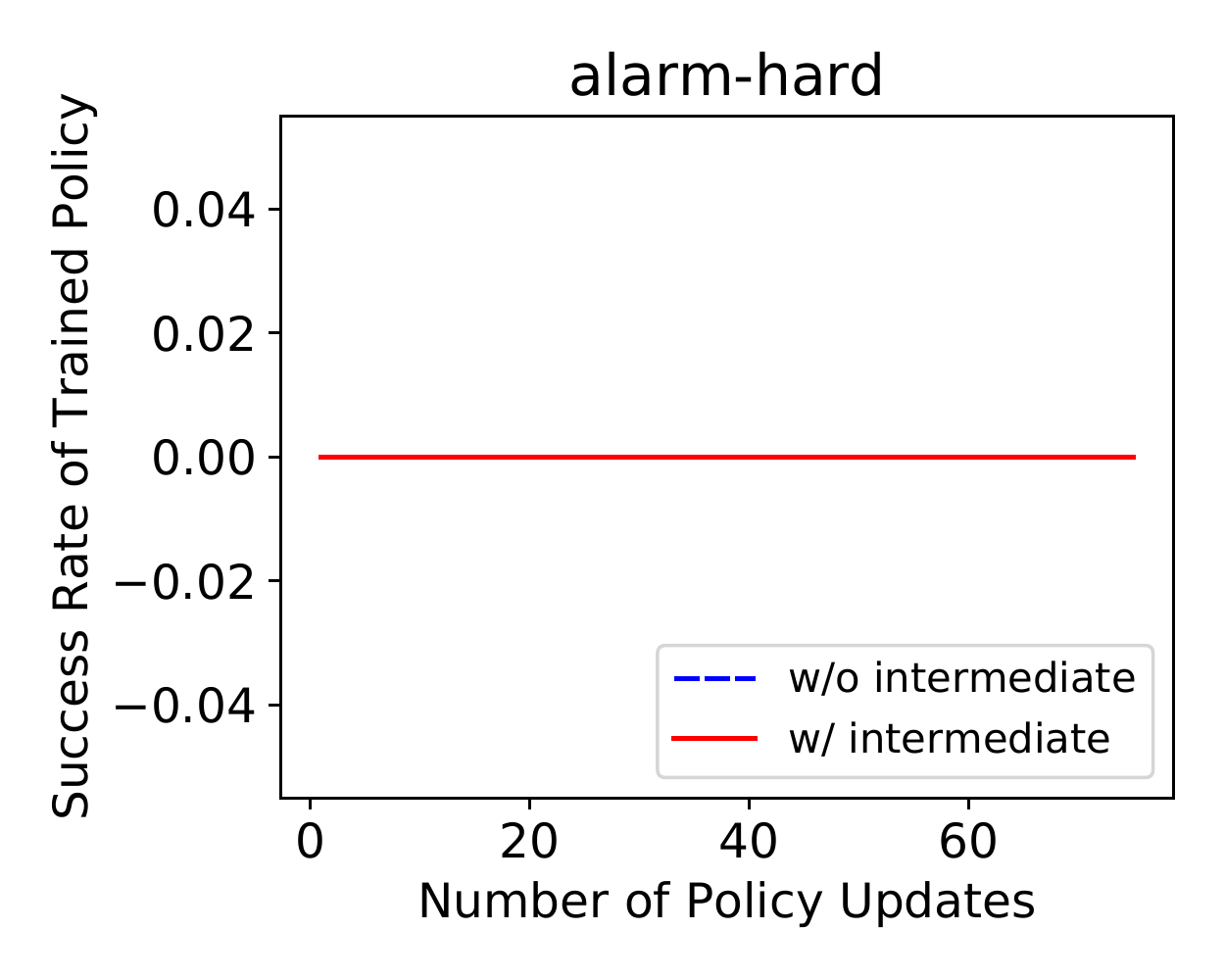}
	}
	
	\caption{The success rate of a trained policy (percentage of successful test episodes) for different reward specification methods in various tasks. \textbf{Comparison between a reward specification with and without intermediate rewards.}}
	\label{fig:intermediate_final_extra}
	\squeezeup
\end{figure*}

Figures~\ref{fig:3_35_extra}-\ref{fig:intermediate_final_extra} show additional results comparing each two baselines on each task in our suite of benchmarks.


